\documentclass[lettersize,journal]{IEEEtran}
\usepackage{amsmath,amsfonts}
\usepackage{algorithmic}
\usepackage{algorithm}
\usepackage{array}
\usepackage{textcomp}
\usepackage{stfloats}
\usepackage{amsmath}
\usepackage{url}
\usepackage{verbatim}
\usepackage{cite}
\usepackage{graphicx}
\usepackage{subcaption}
\usepackage{color}
\usepackage{float}
\usepackage{placeins}
\usepackage{times}
\usepackage[skip=2pt]{caption}
\makeatletter
\renewcommand{\p@subfigure}{\thefigure} 
\newcommand{\subrefp}[1]{(\subref{#1})} 
\makeatother

\begin{document}

\title{Measure the Sim-to-Real Gap: Designing an Affordable Real-World Benchmark Platform for Reinforcement Learning in AIoT Systems}

\author{Rongping Zhou, Omid Tavallaie, Shuaijun Chen, Albert Y. Zomaya
\thanks{Rongping Zhou, Shuaijun Chen, Albert Y. Zomaya are with the School of Computer Science, The University of Sydney, Camperdown NSW 2050, Australia (e-mail: rzho0616@uni.sydney.edu.au; shuaijun.chen@sydney.edu.au; albert.zomaya@sydney.edu.au).

Omid Tavallaie is with the Department of Engineering Science, University of Oxford, Wellington Square, Oxford OX1 2JD, United Kingdom, and the Department of Computer Science, The University of Western Australia, 241, 35 Stirling Hwy, Crawley WA 6009 (email: omid.tavallaie@eng.ox.ac.uk; omid.tavallaie@uwa.edu.au).
}}


\maketitle

\begin{abstract}
Reinforcement learning (RL) is commonly employed to enhance the performance of autonomous systems, including the Autonomous Internet of Things (AIoT). However, the trial-and-error nature of RL, when conducted in real-world environments, is costly and hazardous in some scenarios. Consequently, the majority of RL research is conducted in simulation. This reliance introduces challenges related to the Sim-to-Real transferability. Evaluating the Sim-to-Real algorithmic robustness and the Sim-to-Real gap is a critical prerequisite for research aimed at improving RL performance in the real world. Therefore, industries such as robotics have developed concurrent simulation and physical platforms to facilitate this research. However, a universal Sim-to-Real benchmark platform for AIoT does not currently exist. To address these concerns, we developed a real-world AIoT platform for studying RL in AIoT. On this platform, an agent deployed on an edge device plays video games on a separate host computer via a hardware-emulated keyboard, guided by vision input. This platform uses commercially available components costing less than USD 400, together with two computers. Because the system's objective is game score maximization, it inherently mitigates safety risks associated with real-world RL deployments. Experimental results show the simulation-trained agent suffers a 1160\% performance degradation relative to the human-level performance after real-world deployment, indicating a significant Sim-to-Real gap. Direct real-world training using the deep Q-network (DQN) RL algorithm achieves approximately 49\% of human-level performance after 10 million training steps, demonstrating the feasibility of RL under real-world conditions. These results suggest that the proposed Sim-to-Real benchmark platform provides a substantial foundation for qualitative and quantitative evaluations of RL in real-world AIoT systems.

\end{abstract}


\section{Introduction}
\label{sec:intro}

\IEEEPARstart{R}{einforcement} learning (RL) \cite{RN59}, when combined with deep learning \cite{Goodfellow-et-al-2016}, serves as an effective methodology for improving the performance of autonomous systems, including robotic platforms \cite{doi:10.1177/0278364919887447, pmlr-v87-kalashnikov18a}, game-playing agents \cite{doi:10.1126/science.aar6404}, artificial intelligence systems incorporating Large Language Models (LLMs) \cite{NEURIPS2022_b1efde53}, specialized systems such as tokamak equipment \cite{RN85}, and the AIoT \cite{RN51,RN167,7123563,9069178}. The iterative trial-and-error mechanism inherent in RL can require substantial resources and may introduce safety hazards that can reach a life-threatening level in certain real-world scenarios. As a result, research in these domains employs simulated environments, which provide cost-effective, controlled, and repeatable conditions for RL algorithms, agent development, and evaluation. RL algorithms and agents are typically designed, trained, and tested in simulation before being deployed to real-world environments. RL training is expensive in terms of computing resources, regardless of whether it is conducted in simulation or the real world. The Arcade Learning Environment (ALE) has served as a benchmark platform in RL research since Google DeepMind achieved human-level performance on Atari games with an RL agent in 2015 \cite{RN91}. On this platform, training an RL agent for 200 million frames is a standard approach to achieving results comparable to the original breakthrough performance of the RL algorithm \cite{machado2018revisiting}. This training process demands significant computational resources, requiring days in simulation and months in real-world settings. Given that real-world game execution typically operates at 60 frames per second (FPS), processing 200 million frames would take more than 38 days, excluding additional computational time consumed by the training procedure. This disparity indicates that the computational cost of RL training is considerably higher in real-world environments than in simulation. Consequently, a realistic approach is to train RL agents in simulation, followed by evaluation in the real world. Even for testing purposes, RL testing requires a significant investment in equipment, infrastructure, and physical space. For instance, evaluating an RL agent designed for a data center cooling system requires a spacious facility equipped with costly servers and an expensive cooling system\cite{dccooling}. Furthermore, training and testing RL agents in the real world pose safety hazards that can become life-threatening. In the power grid domain, for example, an RL agent does not have the knowledge of the correct breaker on-off delay times. If the RL agent attempts to toggle multiple breakers in a short interval without accounting for the longer delays inherent in real-world systems, a short circuit can occur in the grid. Such an event may lead to a fire or a system-wide power shutdown, thereby triggering life-threatening situations and causing significant damage.

\begin{figure*}[t]
\centerline{\includegraphics[width=\linewidth]{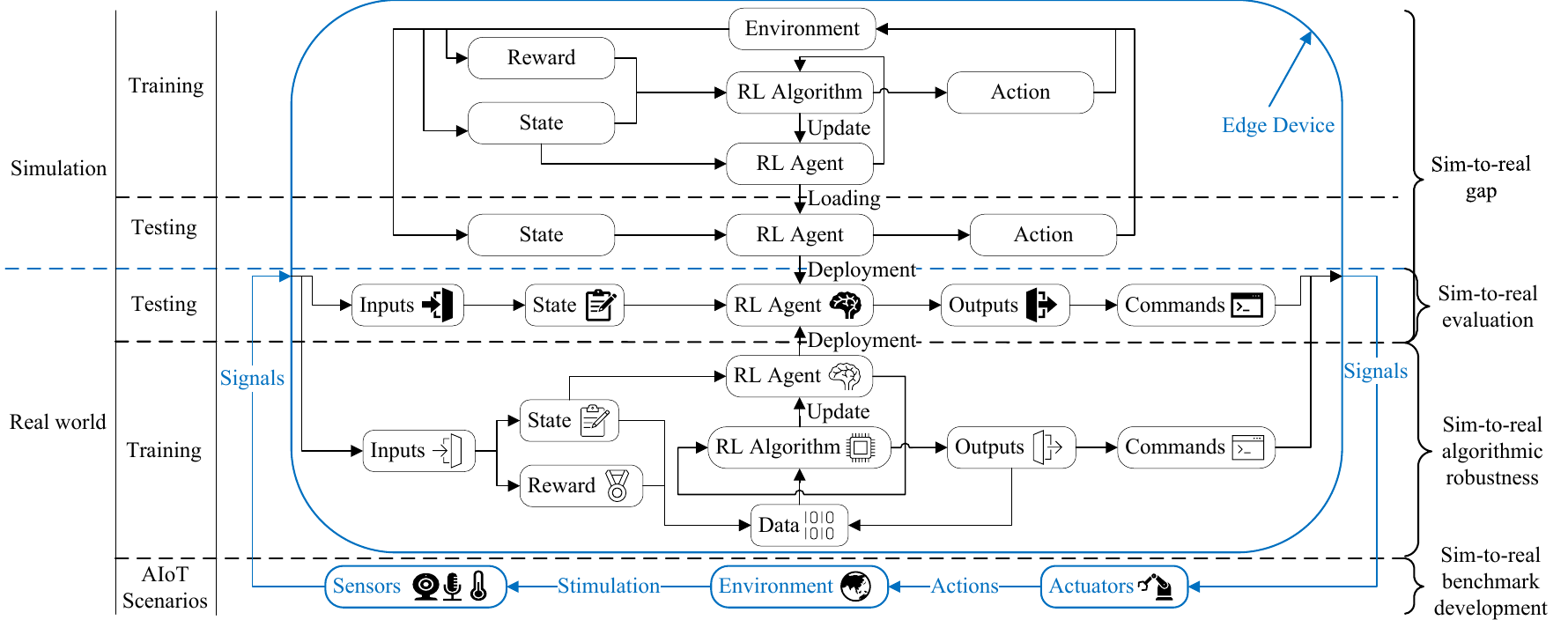}}
\caption{The agent-environment interaction in an AIoT system.}
\label{rlinteractions}
\end{figure*}

From these examples, it is clear that most reinforcement learning research, whether in the AIoT field or other areas, is conducted in simulated environments, and researchers report real-world results infrequently \cite{RN85, 10.1145/3656043, 9032207, 9551113, 9200541, 8919976, 9069178, 9403369}.  Consequently, several challenges arise concerning the Sim-to-Real gap and algorithmic robustness when transferring RL to real-world scenarios, as shown in Fig.~\ref{rlinteractions}. First, it is uncertain whether RL algorithms are applicable to real-world applications despite their perfect performance in simulation. Second, it is necessary to verify the operational viability of simulation-trained agents within noisy physical environments. Finally, it is necessary to develop a Sim-to-Real benchmark platform to evaluate the performance discrepancy between these two domains and precisely measure the Sim-to-Real gap.

In certain domains, including robotics, tokamak equipment, and similar systems, where real-world systems exist, researchers may first investigate the feasibility of RL applications in these fields and subsequently develop a simulation environment to support that effort. For other researchers seeking to extend or reproduce such work, the available options are either to invest significant resources in constructing equivalent systems or obtain access to existing ones. This circumstance significantly limits the reusability of research results produced by the researchers in those domains. In practice, AIoT systems exhibit considerable diversity, including applications such as environmental control systems in smart buildings, traffic control systems in transportation networks, and power distribution in power grids. Most researchers rely on data collected from these domains to construct simulated environments. However, validating RL algorithms or agents in real-world systems remains challenging, as securing sufficient funding to build or access real-world systems equipped with expensive devices is difficult for many research groups. Conducting real-world experiments that yield more accurate results than simulations without incurring substantial costs remains a significant methodological hurdle for RL applications in AIoT systems. To address this issue, the following sections review the architecture of AIoT systems and how RL is applied in AIoT systems.

In the RL paradigm for AIoT systems, as shown in Fig.~\ref{rlinteractions}, AIoT systems consist of sensors, actuators, and edge devices, \cite{RN59,7123563,edge}. Sensors are devices that detect environmental changes, such as cameras. Actuators are devices that change the environment, for example, relays and motors. Edge devices are computing devices that host RL agents and algorithms, receive sensory data, and transmit commands to actuators. These sensors, actuators, and edge devices connect via a variety of communication technologies, including Ethernet and Bluetooth. Within this architecture, the RL agent on an edge device tries to maximize the reward it receives via a trial-and-error process. Firstly, sensors receive environmental stimuli and convert them into inputs for edge devices. Next, the RL agent and algorithm on the edge device interpret these inputs as the state and reward, and subsequently generate commands. After that, the edge device transmits commands to actuators via signals. Then the actuators take actions according to the commands they receive. Finally, the resulting actions change the environment, which in turn stimulates the sensors differently, thereby changing the state and reward the agent receives. The RL algorithm enables the agent to explore diverse actions in response to the current state at each time step, and those actions cause changes in the environment, thereby affecting the rewards the agent receives in subsequent steps. When RL algorithms operate, the trial-and-error interactions make the agent progressively enhance its ability to maximize the long-term cumulative rewards \cite{RN59}.

Additionally, simulation can be executed directly on edge devices. The RL algorithm receives the reward and the state from the simulated environment, processes the input from the simulated environment, generates actions, and transmits actions to the simulated environment. The RL agent's performance improves as the algorithm updates it in the simulation. Following training and testing in the simulated environment, the RL agent can be deployed to the real-world environment for further testing. Then it is natural to examine the difference between the performance of the RL agent in simulation and that in the real world, which is the Sim-to-Real gap measurement problem. As there are differences between simulation and reality, researchers always need to check whether the RL algorithm can work effectively in the real-world environment, which is the Sim-to-Real algorithmic robustness problem. To address these two problems, a Sim-to-Real benchmark platform and a Sim-to-Real evaluation procedure need to be developed.

Based on the preceding examination of AIoT systems and RL applications within them, a functional AIoT system can be constructed using sensors, actuators, and edge devices. A minimal instance of an AIoT system may consist of a simple network connecting an edge device, a sensor and an actuator. From this elementary network, an AIoT system can be expanded into a more complex structure incorporating multiple components, ranging from dozens to even millions of sensors, actuators, and edge devices. Building on the minimal architecture, an affordable, readily accessible real-world evaluation platform for RL in AIoT environments was developed. Playing video games is a standard practice in RL research; researchers can quickly gain experience, and there are no safety concerns when developing a real-world system for playing video games. We used cameras, physical key pressers, a hardware-emulated keyboard and Atari games to build a real-world system. This real-world system enables researchers to further study high-performance algorithms and develop new approaches to improve RL agents in real-world settings. The main contributions are summarized below:
\begin{itemize}
    \item We introduce a platform that facilitates RL training and evaluation directly in real‑world settings. The majority of current RL research relies on simulated environments, primarily to avoid the substantial costs and safety hazards inherent in real-world experimentation.
    \item We propose methods to evaluate the Sim-to-Real gap of RL agents and the Sim-to-Real algorithmic robustness. This is particularly important because existing real-world AIoT systems differ substantially, making it difficult for researchers to compare results across studies. 
    \item Our platform offers a low‑cost implementation, remaining under USD 400, excluding the computers, and all required components can be conveniently purchased online. This affordability contrasts sharply with other real‑world systems, which often require substantial investment, extensive labor, and considerable resources.
\end{itemize}

The paper is organized into nine sections. Section~\ref{sec:intro} is the introduction of this research. Section~\ref{sec:work} presents prior research on RL in real-world scenarios and RL for AIoT. Section~\ref{sec:problem} presents the research problems for this paper. Section~\ref{sec:imp} describes the implementation of the real-world system. Section~\ref{sec:eval} examines the evaluation results on this system. Section~\ref{sec:dis} discusses what these results mean. Section~\ref{sec:con} concludes this paper. Section~\ref{sec:future} presents the future work on RL for real-world AIoT. Section~\ref{sec:platform} covers the availability of this platform's resources.

\section{Related Work}
\label{sec:work}
\subsection{Real-world reinforcement learning is expensive}

\begin{table*}[t]
\centering
\caption{Comparison between reinforcement learning applications}
\label{tab:rl_comparison}
\renewcommand{\arraystretch}{1.18}
\setlength{\tabcolsep}{3.2pt}
\footnotesize
\begin{tabular}{p{0.135\linewidth} c c c p{0.155\linewidth} p{0.155\linewidth} p{0.155\linewidth} p{0.155\linewidth}}
\hline
\textbf{System} 
& \textbf{IoT} 
& \textbf{Sim.} 
& \textbf{Real} 
& \textbf{Cost} 
& \textbf{Reward} 
& \textbf{Benchmark} 
& \textbf{Baseline} \\
\hline

Atari games \cite{brockman2016openaigym, towers2024gymnasium, 10.5555/2566972.2566979, farebrother2024cale, javed2026physicalatarirobustaccessible}
& \textasciitilde
& \checkmark
& \checkmark
& Low \newline as the system reported in \cite{javed2026physicalatarirobustaccessible} was unavailable at the commencement of this research, a dedicated benchmark platform was independently designed and implemented to support this project
& Game scores
& The trained agent plays the Atari games in the Gymnasium/ALE simulation environment or on the real-world platform, and then compares the performance with that of a human player
& Average human player or the best human player \\

\hline

Robots \cite{6386109,RN248,menagerie2022github,RN137,coumans2021,mittal2023orbit}
& \textasciitilde
& \checkmark
& \checkmark
& High \newline buy commercially available robots or build customized robots
& Specific reward functions designed for different robot tasks, such as walking, grasping, etc.
& The trained agents conduct tasks in the simulation environments such as MuJoCo, PyBullet and Isaac Sim, or on the real robot hardware
& SOTA performance \\

\hline

Robotic hand \cite{doi:10.1177/0278364919887447}
& $\times$
& \checkmark
& \checkmark
& High \newline need to set up a complex system with cameras and sensors, and the system is maintained by a team of 16 people
& A scalar reward derived from the rotation angles of the hand
& Test procedure inside this real-world environment
& The team works out their own baseline for this system \\

\hline

Seven-robot system \cite{pmlr-v87-kalashnikov18a}
& $\times$
& \checkmark
& \checkmark
& High \newline need to set up a lab with seven robots and the items for grasping tasks
& A scalar reward derived from the successful grasp results
& Test procedure inside this real-world environment
& Supervised learning method \\

\hline

Tokamak equipment \cite{RN85}
& \checkmark
& \checkmark
& \checkmark
& High \newline need to get access to expensive tokamak equipment that requires a lot of resources 
& Single scalar reward that is calculated by combining multiple objectives, such as plasma current, radius of the plasma, etc.
& Specific testing procedure conducted on the real-world tokamak equipment 
& Previous experimental results with methods other than RL \\

\hline

IoT-based river flow management system \cite{9551113}
& \checkmark
& \checkmark
& $\times$
& High \newline need to get access to a river where the construction and modification of groynes can be conducted.
& A specific reward based on the river health index
& Test procedure inside the simulation environment
& The river health index before applying the RL method \\

\hline

Smart building \cite{10.1145/3656043, 8919976, 9403369}
& \checkmark
& \checkmark
& $\times$
& High \newline need to get access to a building with extensive building automation equipment 
& Specific reward functions that consider targets for smart buildings, such as energy savings, occupant comfort, etc.
& Test procedure in the simulation environment
& The results from the methods other than the RL method, or the results achieved during research  \\

\hline

Smart transportation \cite{9032207, 9403369}
& \checkmark
& \checkmark
& $\times$
& High \newline need to get access to a transportation system with a spacious area, equipment, and devices for traffic control
& Specific reward functions considering targets for smart transportation, such as waiting time at intersections.
& Test procedure inside the simulation environment
& The results from previously tested RL approaches \\

\hline

\end{tabular}

\vspace{1mm}
\begin{flushleft}
\footnotesize
\textit{Note:} IoT = The system is an IoT system; Sim. = The evaluation is done in simulation; Real = The evaluation is conducted in the real world; Cost = The cost of conducting the evaluation in the real world; Reward = How reward is defined in this system; Benchmark = The standardized test for this system; Baseline = A reference for comparison in this system; it could be an experimental result, a method, or a model.
$\checkmark$ indicates available, $\times$ indicates not available, and\;\textasciitilde\; indicates not applicable.
\end{flushleft}
\vspace{-2mm}
\end{table*}

Real-world RL experimentation entails substantial financial and operational costs across diverse applications, including nuclear fusion facilities \cite{doi:10.1177/0278364919887447}, robotics \cite{RN85}, and other complex real-world infrastructures.

DeepMind and the Swiss Plasma Centre collaborated to develop an RL model to solve the tokamak magnetic control problem in a nuclear fusion experiment \cite{RN85}. Tokamak equipment is considered IoT equipment as actuators are used to control the plasma shape based on the input from the sensors. The model processed inputs such as magnetic flux, magnetic field, and controller parameters, and produced reference voltage commands for the control coils. Training was conducted in a tokamak simulator, after which the architecture was evaluated on real tokamak equipment. The reported results indicated that the learned control method could manage complex tasks and respond effectively to previously unknown situations. This system has its own benchmark, and the baseline is based on experimental results from methods other than RL. However, the capacity for further study remains limited to institutions with access to this nuclear fusion device.

A group of researchers at OpenAI demonstrated the use of RL for controlling a physical Shadow Dexterous Hand \cite{doi:10.1177/0278364919887447}, a commercially available robotic hand with 5 fingers and 24 degrees of freedom. The hand is controlled by electric motor actuators, whose control signals were represented as a 20-dimensional action space corresponding to joint-angle commands. The experimental platform incorporated a highly instrumented physical setup, including 16 PhaseSpace tracking cameras, 3 Basler RGB cameras, and 26 Hall-effect sensors. Visual data from cameras and sensor data were integrated into a three-dimensional tracking and pose-estimation pipeline, and the resulting measurements were encoded into a 60-dimensional state representation containing angular, rotational, and velocity information. In addition to hardware integration, the study required careful design of rewards, timing and the associated simulation environment. With the efforts of a 16-person team, the project demonstrated that RL can increase the dexterity of the robot hand. At the same time, it illustrated the substantial resources required to construct and maintain a real-world research environment. The benchmark depends on the scale and specificity of the real-world system, which also constrains reproducibility, because subsequent benchmarking by other researchers would require access to an equivalent setup.

A joint team from Google and UC Berkeley constructed a real-world platform comprising 7 robots \cite{pmlr-v87-kalashnikov18a}. The task in this setting was to enable these robots to grasp objects from the containers in front of them and place them in the bins beside those containers, using shoulder-mounted camera input as the primary perceptual signal. The researchers developed a new RL algorithm to train a model from 580K real-world grasps performed by these 7 robots. During evaluation, the trained RL agent achieved a 96\% success rate in grasping objects present during training. Although the study demonstrated the feasibility of large-scale real-world learning without using a simulator, it also underscored the considerable expense of establishing and maintaining a multi-robot laboratory environment.

In summary, real-world RL research remains expensive to conduct due to the substantial costs of specialized equipment, physical infrastructure, and ongoing maintenance. In practice, some research groups construct simulation environments based on real-world settings to improve training efficiency by enabling faster data collection and more iterative experimentation. Even with simulators, real-world experimentation remains indispensable for assessing the Sim-to-Real gap and evaluating the robustness of reinforcement learning algorithms under physical operating conditions. Despite this necessity, real-world reinforcement learning research continues to impose considerable financial burdens because it depends on specialized equipment, dedicated infrastructure, and sustained maintenance. These constraints also reduce accessibility for the wider research community, as follow-up studies typically require either the construction of comparable systems or direct access to the original facilities. In addition, benchmarking protocols are usually tailored to specific application contexts, which complicates independent reproduction and external validation. Taken together, these factors restrict wider participation in real-world reinforcement learning research and impede meaningful comparison across studies.

\subsection{Most reinforcement learning for IoT is done in simulation}

Reinforcement learning has already been explored in many IoT applications; however, achieving practical deployment and ensuring reproducibility remain challenging in domains such as building automation, traffic management, railway systems, and river flow regulation. The majority of existing studies rely on simulation-based evaluation and have not progressed to deployment in physical environments.

In a smart building case, the researchers formulated an optimization objective to reduce energy consumption while maintaining occupant comfort \cite{10.1145/3656043}. This required the construction of a reward function that captured the trade-off between efficiency and habitability, along with a state representation incorporating variables such as indoor and outdoor temperature, solar irradiance, wind conditions, blind angle, and relative humidity. Action design must also cover multiple subsystems, including HVAC, lighting, blinds, and windows. Because the resulting action space was extremely large (2,371,842 possible actions), substantial architectural exploration was necessary, leading in this case to the adoption of a branching duelling Q-network for value approximation. Although the authors developed a simulation platform based on a decade of weather observations from two cities, evaluation was limited to simulated environments and did not extend to deployment in an operational building. This limitation arose from the practical difficulty of securing an appropriate facility along with the historical data required for real-world implementation. Yu et al. reported similar limitations in a smart home study, where an RL–based HVAC control approach improved energy efficiency in simulation despite the absence of an explicit thermal dynamics model; however, validation was restricted to a custom simulator, with no real‑world deployment demonstrated \cite{8919976}.

Another study proposed a broad RL method for IoT applications that demand rapid response \cite{9032207}. Evaluation was performed using the Simulation of Urban Mobility (SUMO) platform, in which the control objective was to minimize vehicle waiting time at intersections through appropriate reward design and model development. The reported findings suggest that the proposed method, which relies on a simpler structural design, can outperform previous RL approaches in efficiency. However, these conclusions remain limited to simulation, as the algorithm was not implemented in actual traffic intersections.

In another project, researchers designed an Industrial Internet of Things system based on Virtually Coupled Train Sets and introduced an RL control algorithm intended to improve operational efficiency and railway safety  \cite{9200541}. The system integrated multiple sensing modalities, including lidar, radar, and camera data, while train-state information was exchanged through WLAN and LTE-M communication channels. Train separation was managed through wireless control mechanisms. Although the proposed cooperative control algorithm achieved strong performance in simulation with respect to both safety and efficiency, its evaluation did not extend to operational railway environments. Consequently, further research by other groups would depend on access to the original simulator rather than a shared real-world benchmark.

Liu et al. investigated an IoT-based river flow management system motivated by the risk of flooding caused by channel meandering associated with riverbed growth and path alteration \cite{9551113}. Their approach sought to regulate meandering through the construction and modification of groynes. The system employed sensors such as cameras and radars to measure river conditions, while control actions were the formation or deformation of groynes, enabling the implementation of an Artificial Variable-Width Channel along the river. Although the researchers built a physical experimental platform for AVWC control, the RL strategy was developed and evaluated solely in a simulation environment. While simulation results indicated promising performance, the lack of validation on the existing real‑world setup limited the empirical robustness of the study.

Two independent survey studies \cite{9069178,9403369} covering applications such as power grids, traffic control, smart factories, connected vehicles, and connected robots similarly concluded that real-world deployment remained a major challenge in IoT-oriented reinforcement learning. Most existing work relies on simulator-generated data, and transferring trained models to operational environments is often impractical. The primary constraints include the large physical footprint of many IoT systems, the extensive equipment required, the high financial burden of experimentation, and the potential safety risks associated with deployment failures.

To summarize, academic research groups often lack the resources to build and maintain full‑scale experimental environments, and even when such resources are available, unsuccessful experiments may cause material damage or pose safety risks. Consequently, the majority of RL studies for IoT systems are conducted in simulated environments.

\subsection{RL Benchmarks and Real-World Evaluation}

Benchmarks for RL research have been developed by multiple research communities, and this section reviews several widely adopted evaluation platforms. OpenAI Gym and its successor Gymnasium, maintained by the Farama Foundation, provide a unified toolkit that integrates a broad range of RL benchmarks through a standardized application programming interface (API) \cite{brockman2016openaigym, towers2024gymnasium}. This framework has become a foundational component of RL research and development by simplifying interactions with environments and algorithm evaluation.

The Arcade Learning Environment, introduced by Bellemare et al. and later extended to continuous action spaces by Farebrother and Castro \cite{10.5555/2566972.2566979}, \cite{farebrother2024cale}, is one of the most influential RL benchmarks. ALE emulates Atari 2600 games, enabling agents to interact with more than 50 environments using raw visual observations. Depending on the setting, the action space consists of either 18 discrete actions or a three‑dimensional continuous space, with game screens rendered at 160×210 resolution using a 128‑color palette. Rewards are typically defined as per‑step score differences. ALE provides a standardized and reproducible framework that facilitates benchmarking, comparison, and replication of RL algorithms across research groups. Notably, it served as the testbed for achieving human‑level control performance \cite{RN91}, with results independently replicated worldwide. Despite its impact, ALE remained entirely simulation‑based and lacked real‑world deployment at the start of this project. The physical Atari system in \cite{javed2026physicalatarirobustaccessible} became available after the benchmark platform developed in this study had been completed. Moreover, this physical Atari system differs from the proposed platform in both scope and implementation strategy. Specifically, the Atari Devbox device is based on a single-board computer that displays the game screen along with QR codes, which are used to transmit reward signals to an agent running on a separate computer via visual input. An autofocus camera monitors the Devbox screen to capture both the game state and the QR code containing the reward information, while several motors control a joystick connected to the Devbox. This design is primarily oriented toward robotic applications. By contrast, the proposed platform is designed to support real-world AIoT studies. The benchmark platform proposed in this study employs three conventional computers, two manually focused and manually adjustable cameras, a hardware-emulated keyboard, and a serial communication link to implement three experimental systems. Because the actuation subsystem contains no moving components, the proposed platform avoids mechanical wear-out and motor overheating issues that may arise in a motor-based actuation system. Detailed descriptions of the proposed platform are provided in Section~\ref{sec:imp}.

Multi-joint dynamics with contact (MuJoCo) \cite{6386109,RN248,menagerie2022github}, a physics engine developed by Todorov et al., has been widely adopted as a benchmark for continuous‑control RL, particularly in robotic manipulation \cite{6386109, RN137}. Its integration into the Gym framework and subsequent open‑source release established MuJoCo as a standard evaluation platform for continuous control RL, especially robotic learning \cite{brockman2016openaigym}. Other simulation engines, including PyBullet and Isaac Sim, developed by Meta and NVIDIA, respectively \cite{coumans2021, mittal2023orbit}, also support robotic reinforcement learning by providing high‑fidelity physics simulation. In addition, benchmark suites such as the Real‑World Reinforcement Learning (RWRL) Suite, RLBench, and Meta‑World focus on structured robotic learning tasks \cite{yu2021metaworldbenchmarkevaluationmultitask, 9001253}. In robotics research and development, simulation tools have evolved beyond abstract models toward digital twins that closely mirror physical systems. These simulation and digital‑twin platforms enable training in virtual environments prior to deployment, supporting systematic investigation of the Sim‑to‑Real gap and improving the Sim‑to‑Real algorithmic robustness. For instance, Wagenmaker et al. trained a policy in MuJoCo and deployed it on a Franka Emika Panda robotic arm, subsequently analyzing the performance discrepancy between simulation and real‑world execution \cite{NEURIPS2024_8fa068ff}.

Overall, a wide range of RL benchmarks is available for academic research, and real‑world evaluation platforms exist, primarily focusing on robotic systems. 

\subsection{Summary}

Real‑world systems for RL research are costly to construct, require substantial resources, and may introduce significant safety risks in practice. Consequently, most RL studies are conducted in simulated settings, and many widely used benchmarks are simulation‑based. Nevertheless, validation in physical environments remains essential, as agents trained solely in simulation cannot be reliably deployed to physical systems without empirical verification. To address this, simulation models are commonly developed to approximate existing physical systems. In robotics, for instance, digital twins, high‑fidelity simulation models that can dynamically replicate physical system behaviour, have played an important role in accelerating recent research progress. In contrast, for RL in IoT systems, the expense and complexity of building real‑world experimental platforms are often prohibitive. In the absence of a shared real‑world benchmark for IoT reinforcement learning, simulation therefore remains the primary evaluation paradigm. All these conclusions are summarized in Table~\ref{tab:rl_comparison}, which is derived from the previous discussions. This context underscores the need for systematic analysis of the Sim‑to‑Real gap, rigorous assessment of the Sim‑to‑Real algorithmic robustness, and principled methodologies for quantifying performance discrepancies between simulated and physical environments.

\section{Problem Statement}
\label{sec:problem}

The related work and introductory discussion indicate that several substantial barriers continue to constrain RL research in real-world AIoT environments \cite{RN85,9069178,10.1145/3656043,8919976,9032207,9200541,9551113,9403369}. The principal obstacles are summarized as follows:

\begin{itemize}
    \item A substantial proportion of prior studies rely exclusively on specialized simulation environments or domain-specific datasets for selected IoT applications. As these results are seldom validated in operational settings, persistent uncertainty remains regarding Sim-to-Real transferability and practical applicability.
    \item Existing studies commonly use heterogeneous benchmarks tailored to specific IoT systems, limiting cross-study comparability. This fragmentation makes it difficult for researchers to interpret, reproduce, and extend methodological advances reported by other groups.    
    \item Real-world evaluation of RL algorithms is costly because it requires an integrated physical infrastructure composed of sensors, actuators, and supporting hardware. In addition to the high cost of deployment and maintenance, such experimentation may introduce severe safety risks, including the possibility of hazardous or life-threatening failures.
\end{itemize}

Addressing these obstacles requires the formulation of several research problems in RL for IoT studies, stated as follows.
\begin{enumerate}
\item The applicability of reinforcement learning algorithms to real-world applications must be examined even when their performance in simulation appears highly effective.
\item The operational validity of reinforcement learning models or agents trained in simulation must be evaluated under real-world conditions.
\item The performance discrepancy between simulation and real-world environments must be quantified through rigorous measurement procedures and suitable benchmark definitions that enable consistent comparison of trained reinforcement learning agents.
\item A cost-effective and broadly accessible real-world IoT platform with minimal safety concerns should be developed so that researchers can use it without specialized domain expertise, together with a corresponding simulation environment to accelerate reinforcement learning training.
\end{enumerate}

\section{Implementation}
\label{sec:imp}

\subsection{Real-world Edge Computing Platform for Reinforcement Learning}

To address the research problems identified for RL in AIoT systems, this study adopts the minimal architectural abstraction introduced in Section~\ref{sec:intro} as the design basis. In its simplest form, an AIoT system comprises an edge device, a sensor, and an actuator connected within a basic operational network. This minimal configuration can be systematically extended to large-scale deployments involving numerous sensors, actuators, and edge devices. Building on this architectural foundation, an affordable and accessible real-world evaluation platform for RL in AIoT environments was developed. Video games were selected as the target application because they are widely used in RL research, provide a familiar experimental setting, and avoid the safety risks commonly associated with physical real-world control tasks. In the implemented platform, the objective is to achieve high-performance gameplay on another computer. The sensing component is a camera, the actuation component is realized through either a physical key presser or a hardware-emulated keyboard, and the edge device is a personal computer interfaced with the camera and the actuation device.

\begin{figure}[htbp]
    \centering
    \begin{subfigure}{0.21\columnwidth}
        \centering
        \includegraphics[width=\linewidth]{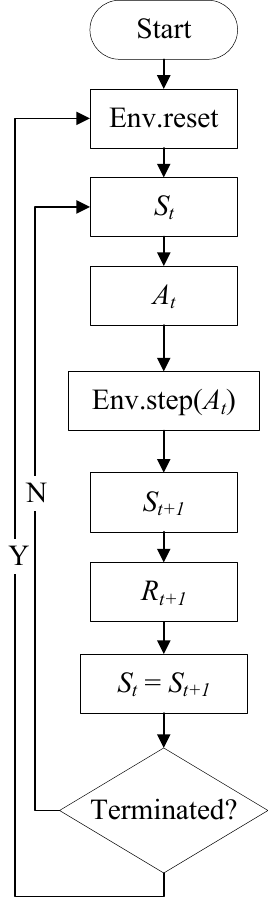}
        \caption{}
        \label{fig:random}
    \end{subfigure}
    \hfill
    \begin{subfigure}{0.38\columnwidth}
        \centering
        \includegraphics[width=\linewidth]{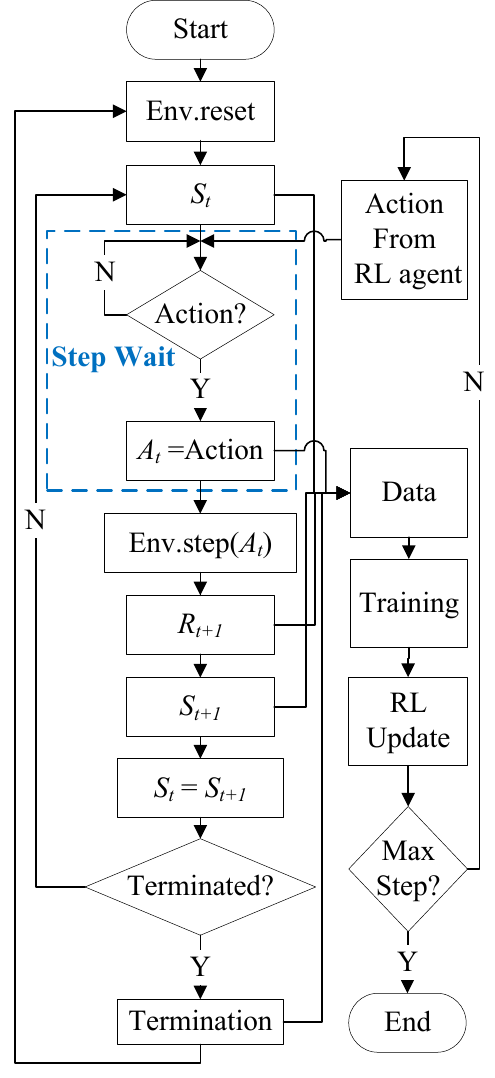}
        \caption{}
        \label{fig:simulation}
    \end{subfigure}
    \hfill
    \begin{subfigure}{0.38\columnwidth}
        \centering
        \includegraphics[width=\linewidth]{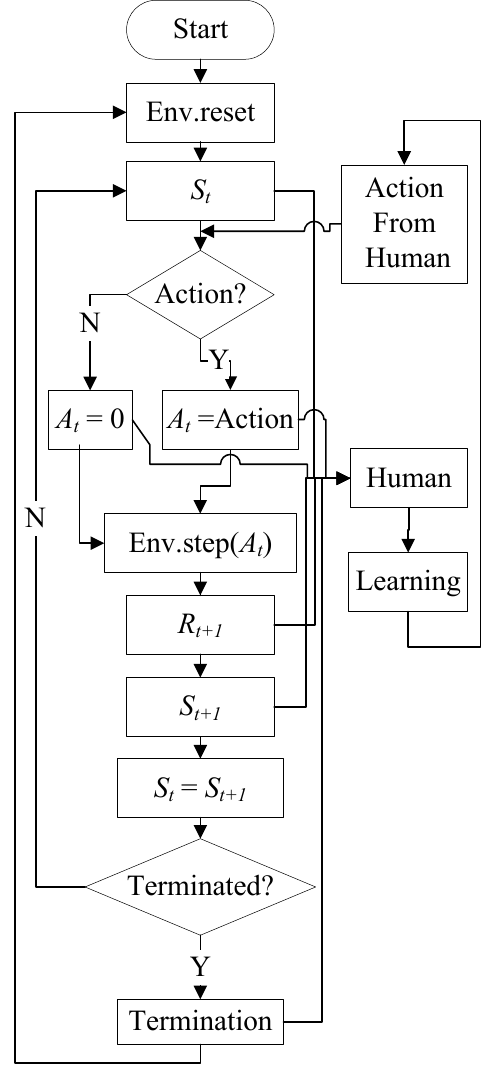}
        \caption{}
        \label{fig:human}
    \end{subfigure}
    \caption{Game playing process. \subrefp{fig:random} Random game playing process. \subrefp{fig:simulation} RL step-wait training process in simulation. \subrefp{fig:human} Human game-playing process; there is no step-wait in the process.}
\label{fig:flow}
\end{figure}

The analysis begins with the interaction flow of the video-game RL environment as illustrated in Fig.~\ref{fig:flow}\subrefp{fig:random}, the corresponding simulation-based training process as shown in Fig.~\ref{fig:flow}\subrefp{fig:simulation}, and the gameplay process executed by a human operator as illustrated in Fig.~\ref{fig:flow}\subrefp{fig:human}.

In the video game RL environment, system execution follows a Markov decision process in Fig.~\ref{fig:flow}\subrefp{fig:random}. Each episode begins with an environment reset, after which the agent repeatedly interacts with the environment until termination occurs. At each time step, the environment produces a state $S_t$, the agent responds with an action $A_t$, and the environment executes the corresponding step function. A reward $R_{t+1}$ is then returned, and the environment advances to the next state $S_{t+1}$. Repetition of this cycle generates a trajectory of $S_t, A_t, R_{t+1},\cdots, S_{T-2}, A_{T-2}, R_{T-1} S_{T-1}, A_{T-1}, R_T$, which extends from the initial state to the termination state. In this notation, the subscript $T$ of $R_T$ denotes termination, whereas $T-1$ represents the final non-terminal step. After termination, the environment is reset and the episodic process begins again.

In the simulation setting, the interaction loop is augmented with an explicit training stage, as shown in Fig.~\ref{fig:flow}\subrefp{fig:simulation}. Following the observation of each state $S_t$, the environment enters a step-wait phase during which execution is suspended until an action $A_t$ is supplied by the RL procedure, either through random exploration or policy-based selection. Without an available action $A_t$, state progression does not occur. During training, each transition, consisting of the current state $S_t$, selected action $A_t$, resulting reward $R_{t+1}$, subsequent state $S_{t+1}$, and termination status, is recorded as training data. The RL algorithm uses these data to update the agent, after which a new action is generated by the RL algorithm and the process is repeated until the predefined maximum number of steps is reached.

In practical gameplay, video games continue to execute regardless of whether a human player provides an immediate response. Within the RL environment, the absence of an input is interpreted as a default action, as illustrated in Fig.~\ref{fig:flow}\subrefp{fig:human}. Consequently, the step-wait mechanism used in simulation lowers the effective control difficulty by allowing the agent unlimited response time before each transition. This characteristic makes the simulation setting more permissive than the real-time interaction conditions experienced by human players.

Following the preceding analysis, the system was developed incrementally. The first implementation replaced the simulation-generated state representation with camera-based input while retaining the simulation program structure, as illustrated in Fig.~\ref{fig:flow}\subrefp{fig:simulation}. The step-wait phase remained part of the training loop in this configuration. This intermediate setup enabled examination of how real-world sensory input influences both agent training dynamics and final performance outcomes. The corresponding physical realization is shown in Fig.~\ref{realinput}. This configuration is hereafter referred to as the real-world input system.

\begin{figure}[htbp]
\centerline{\includegraphics[width=\linewidth]{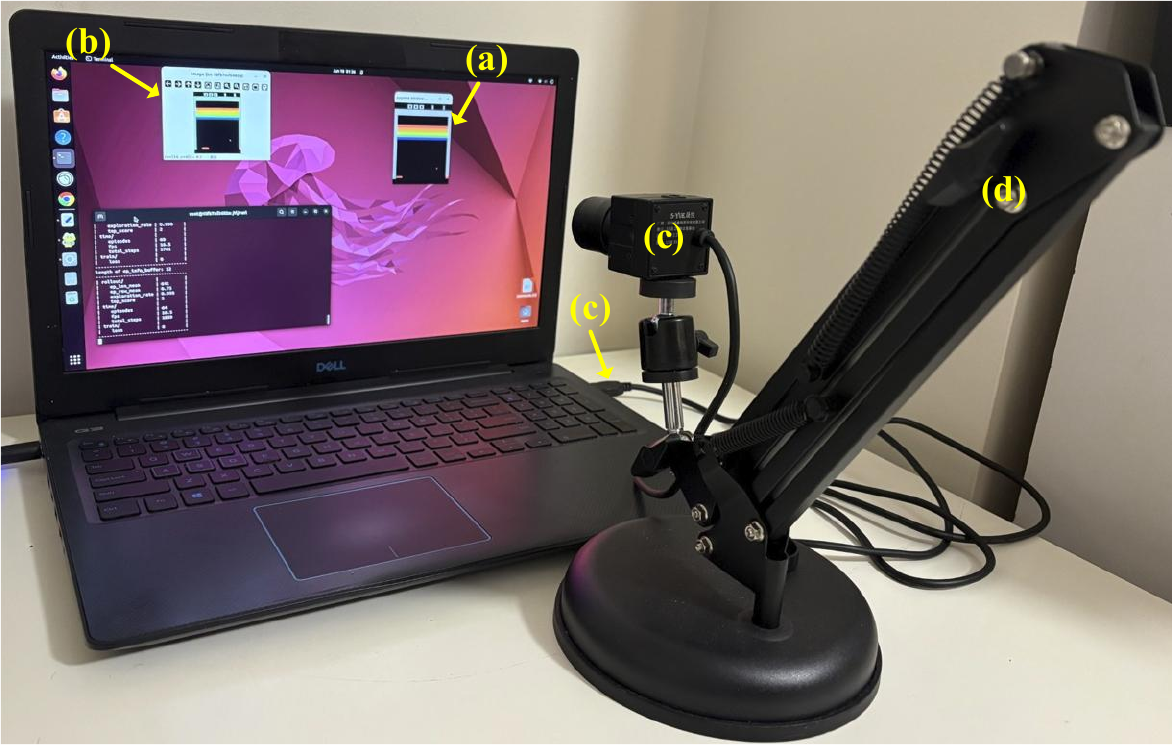}}
\caption{A PC with a camera input playing video games. (a) A game is being played on the computer. (b) Video captured by an RL agent. (c) A camera watching the game is connected to this computer via a USB connection. (d) A camera stand holding the camera.}
\label{realinput}
\end{figure}

\begin{figure}[htbp]
\centerline{\includegraphics[width=\linewidth]{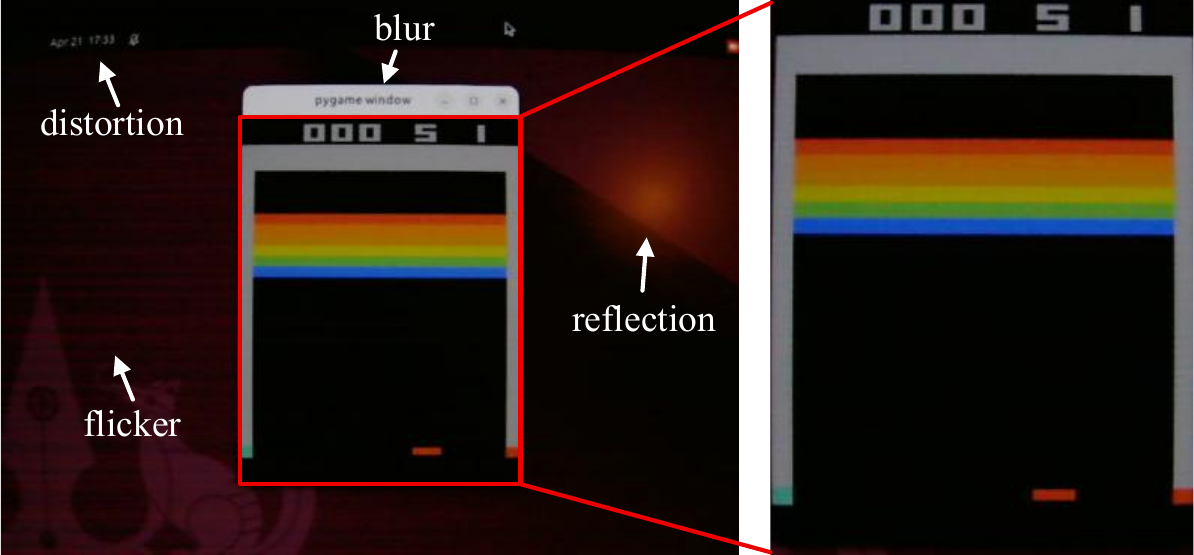}}
\caption{Field of view of the camera. Left: The camera input's original field of view shows noise, including blur, distortion, reflection, and flicker. Right: Cropped field of view of the camera input.}
\label{realinput}
\end{figure}

\begin{algorithm}[!t]\small
\caption{Deep Q-learning (DQN) with experience replay in the real-world input system}
\label{alg:real_input}
\begin{algorithmic}
\STATE Initialize replay memory $D$ with capacity $N$
\STATE Initialize action-value function $Q$ with random weights $\theta$
\STATE Initialize target action-value function $\hat{Q}$ with \\
\hspace{0.1cm} weights $\theta^{-} = \theta$
\STATE Read game settings finite attempt mode into \\
\hspace{0.1cm} variable $has\_lives$
\STATE Initialize camera and point camera toward the \\
\hspace{0.1cm} screen of the computer

\STATE Set $steps = 0$

\FOR{episode $= 1$ to $M$}
    \STATE \textsc{Env.Reset}()
    \STATE Display the environment state on the screen
    \STATE Capture initial image $x_{1}$ from the screen of the computer via the camera, set $s_{1} = x_{1}$ and process sequence $\phi_{1} = \phi(s_{1})$
    \IF{$\text{has\_lives} = 1$}
        \STATE Observe $lives$ from the environment and \\
        set $lives\_after \leftarrow lives$
    \ENDIF
    \FOR{$t = 1$ to $T$}
        \STATE With probability $\varepsilon$ select random action $a_{t}$ \\
        otherwise select $a_{t} = \arg\max_{a} Q(\phi(s_{t}), a; \theta)$
        \STATE \textsc{Env.Step($s_t$)}
        \STATE Observe $r_t$, the status of $terminated$ and \\
        $truncated$ from \textsc{Env.Step($s_t$)}
        \STATE Display the state of environment on the screen
        \STATE Capture the image $x_{t+1}$ via camera, set $s_{t+1} = x_{t+1}$ and process sequence $\phi_{t+1} = \phi(s_{t+1})$
        \IF{$\text{has\_lives} = 1$}
            \STATE Observe $lives$ from the environment
            \STATE Set $done \leftarrow (lives < lives\_after) \ \OR\ terminated$
            \STATE Set $lives\_after \leftarrow lives$
        \ELSE
            \STATE Set $done = terminated$
        \ENDIF

        \STATE Store transition $(\phi_{t}, a_{t}, r_{t}, \phi_{t+1}, done)$ in $D$

        \STATE Sample random minibatch of transitions \\ 
        $(\phi_{j}, a_{j}, r_{j}, \phi_{j+1}, done)$ from $D$

        \STATE Compute target
        \STATE \quad $y_{j} = r_{j}$ if episode terminates at step $j+1$
        \STATE \quad $y_{j} = r_{j} + \gamma \max_{a'} \hat{Q}(\phi_{j+1}, a'; \theta^{-})$ otherwise

        \STATE Perform gradient descent on $(y_{j} - Q(\phi_{j}, a_{j}; \theta))^{2}$ \\
        w.r.t. the network parameters $\theta$

        \STATE Every $C$ steps set $\hat{Q} = Q$
        \STATE Set $steps = steps + 1$
        \IF{$done = True$ \ \OR\ $steps > \text{Max\_Steps}$}
            \STATE Set $T = t$
            \STATE Set $M = episode$ if $steps > \text{Max\_Steps}$
        \ENDIF        
    \ENDFOR
\ENDFOR
\end{algorithmic}
\end{algorithm}

\begin{algorithm}[!t]\small
\caption{The process running at the game computer for the scenario of playing a game with two computers in the real-world system (the agent and the game environment)}
\label{alg:real_world_game}
\begin{algorithmic}
\STATE Read game settings finite attempt mode into variable $has\_lives$
\STATE Initialize serial connection
\STATE Initialize environment state $s = s_0$
\STATE Set control frequency $f$
\STATE Set control period $\Delta t = 1/f$
\STATE Send READY signal to the agent computer
\IF{Receive ACK signal from the agent computer}
    \STATE continue
\ELSE
    \STATE exit
\ENDIF
\WHILE{true}
    \STATE $t_{\text{start}} \leftarrow \textsc{CurrentTime}()$
    \STATE Display state $s$ on the screen

    \STATE $k \leftarrow \textsc{GetKeyboardInput}()$

    \IF{$k \neq \text{None}$}
        \STATE $a \leftarrow k$
    \ELSE
        \STATE $a \leftarrow 0$
    \ENDIF

    \STATE $(s', r, terminated, truncated, lives) \leftarrow \textsc{Env.Step}(s)$

    \IF{$has\_lives = 1$}
        \STATE Send $(lives, terminated, truncated, r)$ to the agent computer via serial connection
    \ELSE
        \STATE Send $(terminated, truncated, r)$ to the agent computer via serial connection
    \ENDIF

    \IF{$terminate = 1 \ \OR\ truncated = 1$}
        \STATE $s \leftarrow \textsc{Env.Reset}()$
    \ELSE
        \STATE $s \leftarrow s'$
    \ENDIF

    \STATE $t_{\text{elapsed}} \leftarrow \textsc{CurrentTime}() - t_{\text{start}}$
    \IF{$t_{\text{elapsed}} < \Delta t$}
        \STATE \textsc{Sleep}$(\Delta t - t_{\text{elapsed}})$
    \ENDIF
\ENDWHILE
\STATE {}
\end{algorithmic}
\end{algorithm}

\begin{algorithm}[!t]
\caption{The RL process running at the agent computer (edge device) for the scenario of playing a game with two computers in the real-world system (the agent and the game environment)}
\label{alg:real_world_edge_device}
\begin{algorithmic}\small
\STATE Initialize replay memory $D$ with capacity $N$
\STATE Initialize action-value function $Q$ with random weights $\theta$
\STATE Initialize target action-value function $\hat{Q}$ with \\
\hspace{0.1cm} weights $\theta^{-} = \theta$
\STATE Read game settings finite attempt mode into variable $has\_lives$
\STATE Initialize camera and hardware-emulated keyboard
\STATE Point camera toward the screen of the game environment computer
\STATE Wait for READY signal from the game environment computer
\STATE Send ACK signal after receiving READY signal from \\
\hspace{0.1cm} the game environment computer and start to play game
\STATE Set $steps = 0$
\FOR{episode $= 1$ to $M$}
    \STATE Capture initial image $x_{1}$ from the screen of the game environment computer via the camera, set $s_{1} = x_{1}$ and preprocess sequence $\phi_{1} = \phi(s_{1})$
    \IF{$has\_lives = 1$}
        \STATE Receive $lives$ from the game environment computer via serial connection, set $lives\_after \leftarrow lives$
    \ENDIF
    \FOR{$t = 1$ to $T$}
        \STATE With probability $\varepsilon$ select random action $a_{t}$ \\
        otherwise select $a_{t} = \arg\max_{a} Q(\phi(s_{t}), a; \theta)$
        \STATE Send action $a_{t}$ to the game environment computer via hardware-emulated keyboard
        \IF{$has\_lives = 1$}
            \STATE Receive reward $r_{t}$ and status of terminated, truncated, lives from the game environment computer via serial connection
            \STATE Set $pseudo\_end = False$
            \IF{$lives\_after < lives$}
                \STATE Set $pseudo\_end = True$
            \ENDIF
            \STATE Set $done = pseudo\_end \ \OR\ terminated$
            \STATE Set $lives\_after \leftarrow lives$
        \ELSE
            \STATE Receive reward $r_{t}$ and the status of terminated, truncated from the game environment computer via serial connection
            \STATE Set $done = terminated$
        \ENDIF
        \STATE Capture next image $x_{t+1}$ via camera, set $s_{t+1} = x_{t+1}$ and preprocess $\phi_{t+1} = \phi(s_{t+1})$
        \STATE Store transition $(\phi_{t}, a_{t}, r_{t}, \phi_{t+1}, done)$ in $D$
        \STATE Sample random minibatch of transitions \\ 
        $(\phi_{j}, a_{j}, r_{j}, \phi_{j+1}, done)$ from $D$
        \STATE Compute target
        \STATE \quad $y_{j} = r_{j}$ if episode terminates at step $j+1$
        \STATE \quad $y_{j} = r_{j} + \gamma \max_{a'} \hat{Q}(\phi_{j+1}, a'; \theta^{-})$ otherwise
        \STATE Perform gradient descent on $(y_{j} - Q(\phi_{j}, a_{j}; \theta))^{2}$ \\
        w.r.t. the network parameters $\theta$
        \STATE Every $C$ steps set $\hat{Q} = Q$
        \STATE Set $steps = steps + 1$
        \IF{$done = True$ \ \OR\ $steps > \text{Max\_Steps}$}
            \STATE Set $T = t$
            \STATE Set $M = episode$ if $steps > \text{Max\_Steps}$
        \ENDIF        
    \ENDFOR
\ENDFOR

\end{algorithmic}
\end{algorithm}

Since the camera captures a field of view that extends beyond the active game window, an additional visual recognition module would be required to identify the relevant display region for gameplay from the raw video stream. Incorporating such a module would increase system complexity. To maintain a simpler implementation, the captured frames are cropped directly to the target game window displayed on the monitor, thereby aligning the camera-derived input with the observation format used in the simulation environment. A comparison between the original capture and the cropped result is presented in Fig.~\ref{fieldofview}.

The algorithm for the real-world input system is presented in Algorithm~\ref{alg:real_input}. The proposed procedure follows the deep Q-network (DQN) framework with several modifications required for camera-based real-world input. First, the camera is initialized and aligned with the target display region to ensure consistent acquisition of the area of interest. During execution, the game window remains visible on the screen, unlike in the simulation setting, where on-screen rendering is unnecessary and can reduce training efficiency. At each iteration, the camera acquires visual observations that are processed into a sequence composed of consecutive cropped frames, whereas in the simulation environment, the corresponding sequence is obtained directly from state outputs generated by the game engine. Apart from these input-related adaptations, the remaining components of the procedure are identical to those of the standard DQN algorithm in the simulation environment.

\begin{figure}[htbp]
\centerline{\includegraphics[width=\linewidth]{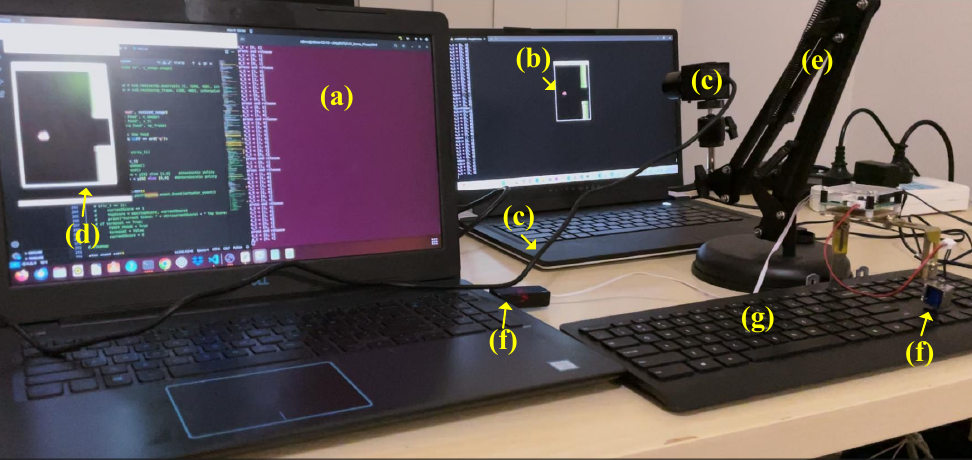}}
\caption{A PC with a camera and an actuator playing video games. (a) An RL agent running on this computer. (b) A game running on another computer. (c) A camera watching the game is connected to the RL agent computer via a USB cable. (d) Captured game video with the camera. (e) A camera stand holding the camera. (f) An actuator presses the key on a wireless keyboard for the game computer; the actuator is connected to the RL agent computer via a USB cable. (g) The wireless keyboard for the game computer.}
\label{physicalpresser}
\end{figure}

The second implementation employed a camera as the sensing device and a physical keyboard presser as the actuation mechanism, as illustrated in Fig.~\ref{physicalpresser}. This configuration introduced two major limitations. First, the number of required physical pressers scales with the number of control actions. In the Flappy Bird task \cite{flappybird, a3c_flappybird}, only two actions are required, namely pressing the up key or taking no action, and therefore a single presser is sufficient. In contrast, the Breakout \cite{breakout} task requires four discrete actions: no key press, pressing the space bar to launch the ball, pressing the right key to move the paddle right, and pressing the left key to move the paddle left. Under this action design, three separate physical pressers are required, which substantially increases hardware complexity. Second, repeated mechanical key pressing generates considerable acoustic noise during operation. To address these limitations, a hardware-emulated keyboard was adopted to replace both the physical presser and the physical keyboard. This modification enables the direct transmission of keyboard commands from the control program to the game computer via the USB interface, without mechanical noise. This modification also falls within the scope of AIoT, as actuation in AIoT focuses on sending the right commands at the right time, whereas robotic applications involve additional considerations, such as motor control algorithms.

\begin{figure}[htbp]
\centering
\includegraphics[width=\linewidth]{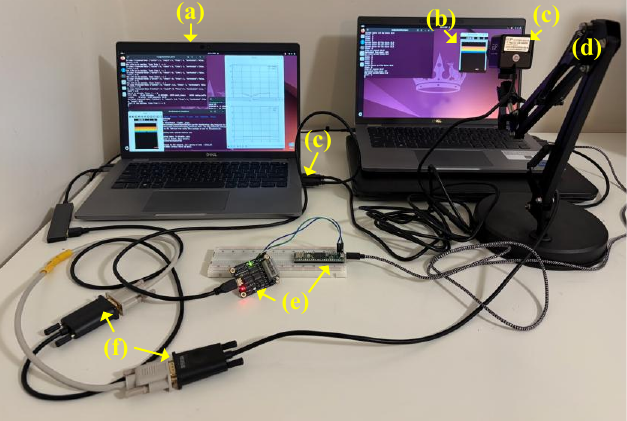}
\caption{A PC with a camera and an emulated keyboard playing video games. (a) An RL agent running on this computer. (b) The game is being played on another computer. (c) A camera watching the game is connected to this RL agent computer via a USB connection. (d) A camera stand holding the camera. (e) A hardware-emulated keyboard made of a USB-to-serial converter, jump wires, breadboard, Teensy board, and USB connections; it receives the command from the RL agent computer and sends the command to the game computer. (f) The serial communication connection between two computers via two USB-to-serial converters.}
\label{realworld}
\end{figure}

A hardware-emulated keyboard based on a Teensy board \cite{teensy41,etherkey} can issue key-press and key-release signals with negligible intrinsic delay, whereas both a human player and a mechanical key presser exhibit nonzero keystroke timing determined by key-press duration and inter-key interval \cite{keystroke}. Consequently, the hardware-emulated keyboard requires an explicitly defined keystroke schedule that reflects realistic human input timing. Because human players can reduce the inter-key interval through key rollover, wherein a subsequent key is depressed before the preceding key is released, the inter-key interval may be treated as negligible in the configuration of the hardware-emulated keyboard. Under this condition, only the key-press duration must be specified. Large-scale keystroke studies have reported that the average key-hold duration is 116.24 ms \cite{keystroke}. Accordingly, the firmware of the hardware-emulated keyboard is configured with a key-press duration of 116.24 ms.

A further system-level challenge involves reward acquisition when the video game is executed on a separate computer. In contrast to simulation environments, where rewards are returned directly by the environment, the physical implementation requires an explicit communication mechanism to transmit score information from the game machine to the edge device. To address this requirement, the video game program was configured to send score and status data to the edge device through a serial communication channel. A direct serial link was then established between the game machine and the edge device, enabling the agent to receive rewards and related execution information in real time. The resulting configuration, illustrated in Fig.~\ref{realworld}, is designated as the real-world system. In the corresponding training workflow, the step-wait phase is removed, and environment transitions proceed continuously regardless of whether an action is executed by the actuator, thereby aligning the interaction process more closely with the human gameplay pattern shown in Fig.~\ref{fig:flow}\subrefp{fig:human}.

The real-world system is implemented through two coordinated algorithms executed on separate computers, which are Algorithm~\ref{alg:real_world_game} on the game machine and Algorithm~\ref{alg:real_world_edge_device} on the agent machine (edge device). In this architecture, the game machine runs the video-game environment, while the agent machine executes the DQN-based control procedure. To support interaction between the two machines, several modifications are introduced. The game machine performs an initial handshaking procedure by transmitting a READY signal to the agent machine and waiting for an ACK message before gameplay begins. To preserve real-time execution conditions comparable to human play, a delay is inserted into each iteration so that the game operates at a standard frame rate, such as 60 FPS. During execution, the game machine also transmits reward values and termination information to the agent machine through a serial communication link. These data provide the information required for subsequent agent training.

On the agent computer, the control procedure follows the DQN framework, with several modifications introduced for real-world operation. A handshaking stage is incorporated to synchronize execution with the game computer before interaction begins. The camera is initialized and aligned with the region of interest on the remote display. During each iteration, a sequence of consecutive camera frames is captured, cropped to the game window, and organized into a sequence that serves as the real-world counterpart to an equally sized sequence of successive state outputs in the simulation environment. The agent computer receives reward values, remaining lives, and termination status from the game machine through the serial communication link. The selected action is then transmitted to the game computer via the hardware-emulated keyboard. Apart from these perception, actuation and communication adaptations, the remaining procedure is consistent with the standard DQN algorithm. Consequently, the edge device plays the game by observing the camera stream, issuing control commands via a hardware-emulated keyboard with a fixed key-press duration, and using the data from the game computer via serial communication and visual input for training.

For the testing procedure, the random action selection component and the DQN training updates are removed from Algorithm~\ref{alg:real_input} and Algorithm~\ref{alg:real_world_edge_device}. Under this configuration, the trained agent is evaluated without further learning during execution.

Following these design steps, three system configurations are defined for comparative evaluation: the real-world system with integrated sensing and actuation, the real-world input system that relies on camera-based observations, and the original simulation environment. These configurations provide a basis for systematic comparison of training behavior and final agent performance under different experimental conditions.

\subsection{Benchmarks and Baselines}

Evaluation in RL research commonly relies on task-specific game metrics, including score, level progression, and remaining lives. In Atari-based studies, human performance records are also widely used as reference baselines, typically represented by average-human and best-human scores. These baselines have become standard points of comparison in RL research.

To enable consistent comparison across agents and settings, this study adopts the Human Normalized Score (HNS) as the primary evaluation metric. The metric normalizes an agent’s score relative to a random-policy baseline and a human-performance baseline, thereby expressing performance on a comparable scale across experiments. The HNS is presented below in Equation~\ref{eqn1}.
\begin{equation}
    \label{eqn1}
    {HNS} = \frac{{Agent}_{score} - {Random}_{score}}{{Human}_{score} - {Random}_{score}}.
\end{equation}

In this equation, ${Random}_{score}$ denotes the performance obtained by an uninformed policy, ${Agent}_{score}$ denotes the performance achieved by the evaluated RL agent, and ${Human}_{score}$ denotes either the average-human or best-human reference, depending on the comparison objective. When the average-human score is used, an HNS value of 1.0 indicates human-level performance. When the best-human score is used, an HNS value greater than 1.0 indicates superhuman performance.

Training for 200 million frames in the Arcade Learning Environment (ALE) is a widely recognized benchmark in deep reinforcement learning \cite{machado2018revisiting}. However, this budget is impractical in the present real-world settings. With an effective execution rate of approximately 60 FPS, a 200-million-frame run would require roughly 38 days, excluding additional optimization overhead. For this reason, the experiments in this study use a training budget of 10 million frames, which remains feasible within the temporal constraints of the real-world platform.

To characterize training performance, the mean score over the most recent 100 episodes is tracked throughout learning. Stable improvement is reflected by a gradual increase in this running average without pronounced oscillations or abrupt performance collapse.

Final performance is evaluated over 100 test episodes, and the resulting scores are converted to HNS using the average-human baseline. From these normalized scores, four summary statistics are reported: mean, median, interquartile mean (IQM), and optimality gap. The IQM is computed as the mean of the central 50\% of the score distribution after discarding the lowest and highest quartiles; the optimality gap is defined as the shortfall from an HNS value of 1.0; for example, an HNS of 0.2 corresponds to an optimality gap of 0.8 \cite{iqm}. Higher mean, median, and IQM values, together with a smaller optimality gap, indicate stronger performance.

\subsection{Benchmark Platform}

The implemented benchmark platform comprises three experimental systems, a standardized benchmark test procedure, Sim-to-Real algorithmic robustness assessment, Sim-to-Real gap measurement, baselines, and the code for this benchmark platform.

The three experimental systems are the real-world system with integrated sensing and actuation, the real-world input system that relies on camera-based observations, and the original simulation environment.

The benchmark test procedure is as follows: The agents are trained for 10 million steps across three systems, and the mean score over the latest 100 episodes is recorded to monitor learning performance. After training, the simulation-trained agent is evaluated for 100 episodes in both the real-world system and the real-world input system, respectively, and the scores for these episodes are recorded. The agents trained in the real-world input system and the real-world system are also evaluated for 100 episodes in their corresponding systems; the scores for these episodes are recorded.

Sim-to-Real algorithmic robustness is assessed by comparing the maximum mean score over the latest 100 training episodes, the learning curve trends across systems, and the human normalized scores converted from the evaluation results of the agents trained in both real-world related systems.

The Sim-to-Real gap is measured using the following method: The scores of all test episodes for the simulation-trained agent in three different systems are converted to human normalized scores according to Equation~\ref{eqn1}. Then the mean, median, interquartile mean (IQM), and optimality gap of each of the three groups of 100 test episodes in three systems are calculated. Then the Sim-to-Real gap is calculated from the difference between the data from testing in simulation and those from the test episodes in the two real-world related systems, respectively.

The baseline is defined by human player performance embedded in the human normalized score index, which provides a reference level for interpreting agent performance relative to human gameplay. 

\section{Evaluations}

\begin{figure}[t]
    \centering
    \begin{subfigure}{0.45\columnwidth}
    	\centering
        \includegraphics[width=\linewidth]{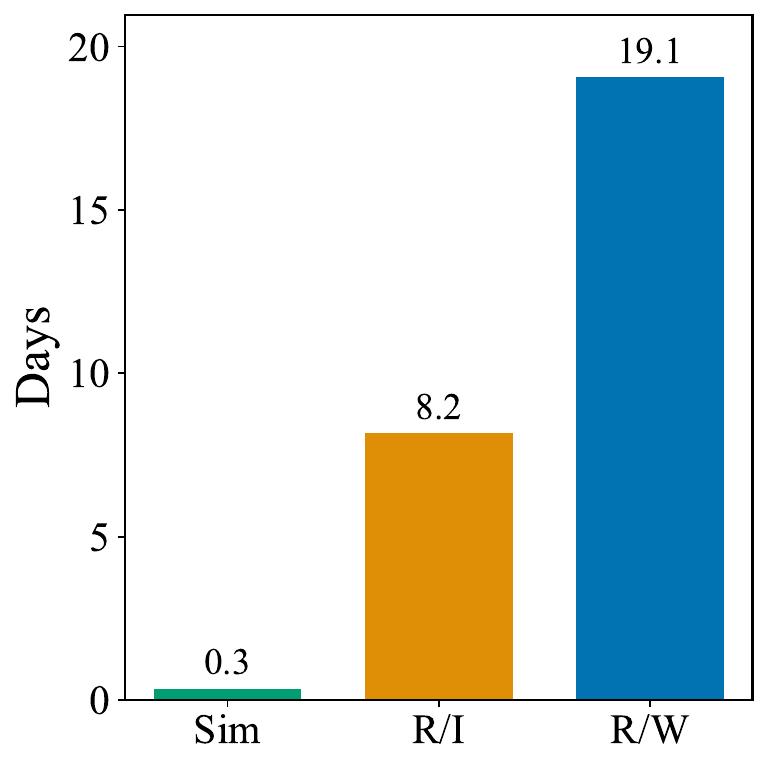}
        \caption{Training time of three agents across three systems}
        \label{fig:training_time_of_three_agents}
    \end{subfigure}
    \hspace{1mm}
    \begin{subfigure}{0.45\columnwidth}
    	\centering
        \includegraphics[width=\linewidth]{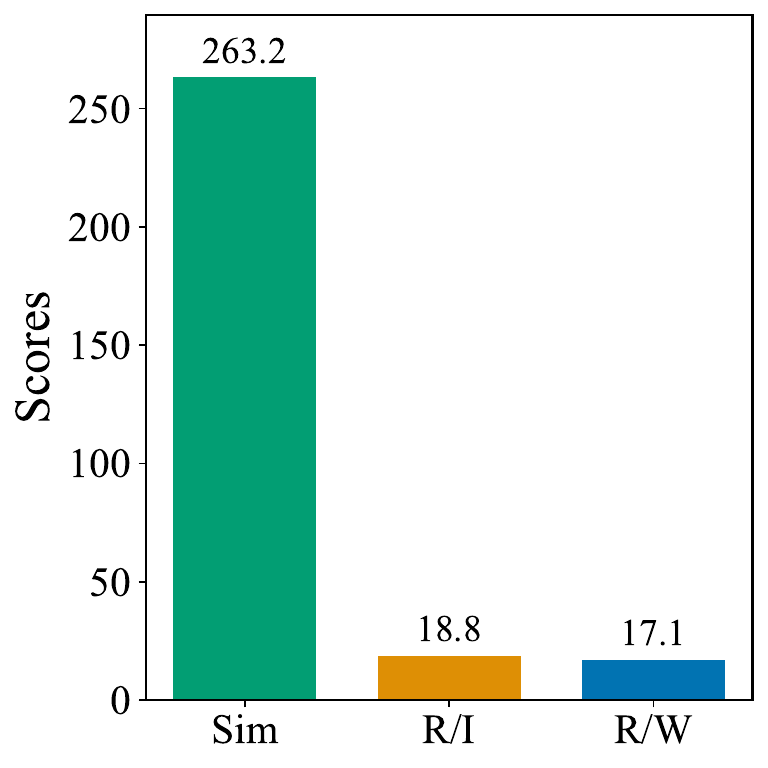}
        \caption{Maximum mean scores of the last 100 training episodes}
        \label{fig:maximum_mean_socre_last_100_episodes}
    \end{subfigure}
    \hspace{1mm}
    \begin{subfigure}{0.45\columnwidth}
       \centering
        \includegraphics[width=\linewidth]{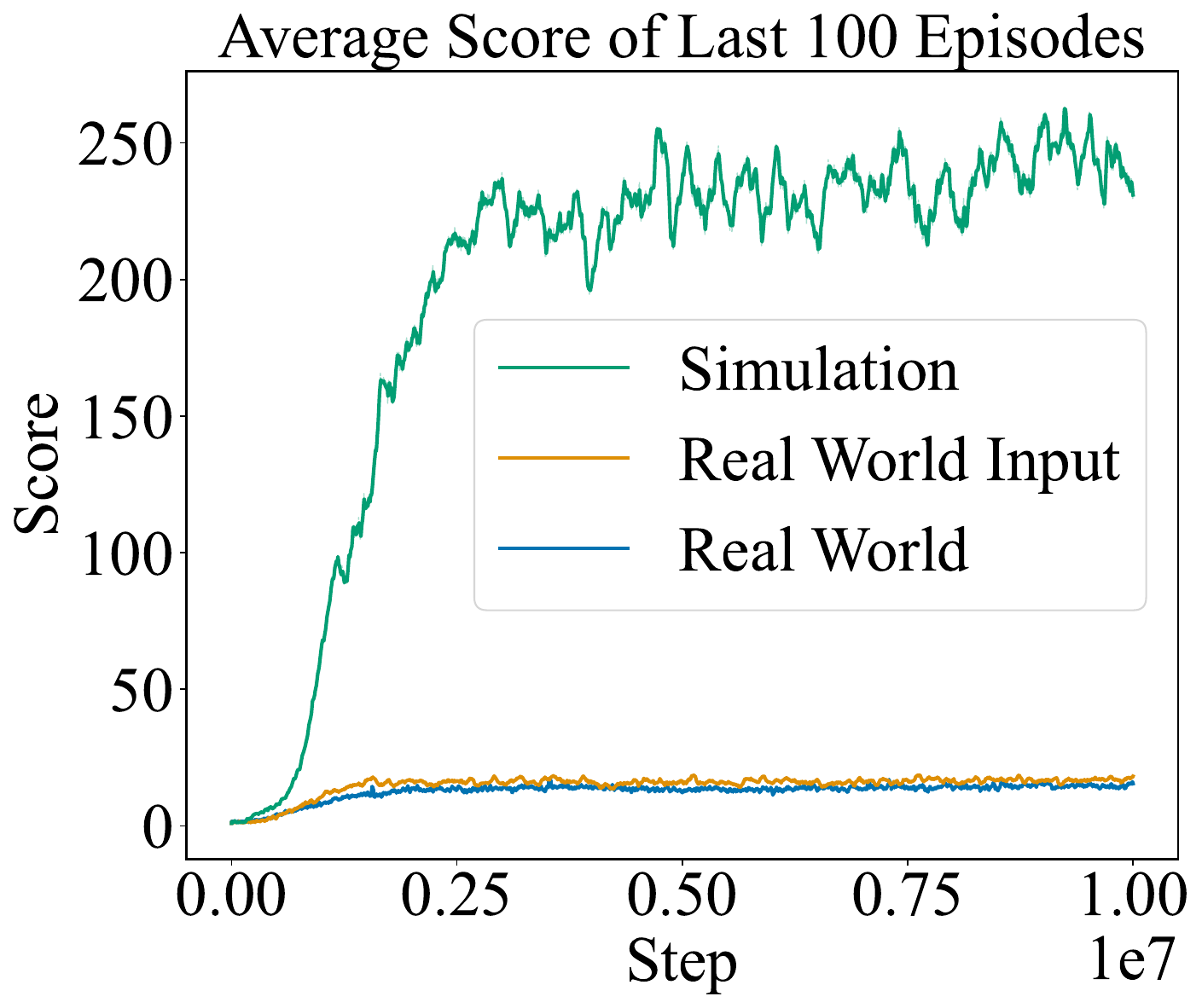}
        \caption{Learning curves of three agents across three systems}
        \label{fig:learning_curve_three_agents_three_systems}
    \end{subfigure}
    \hspace{1mm}
    \begin{subfigure}{0.45\columnwidth}
    	\centering
        \includegraphics[width=\linewidth]{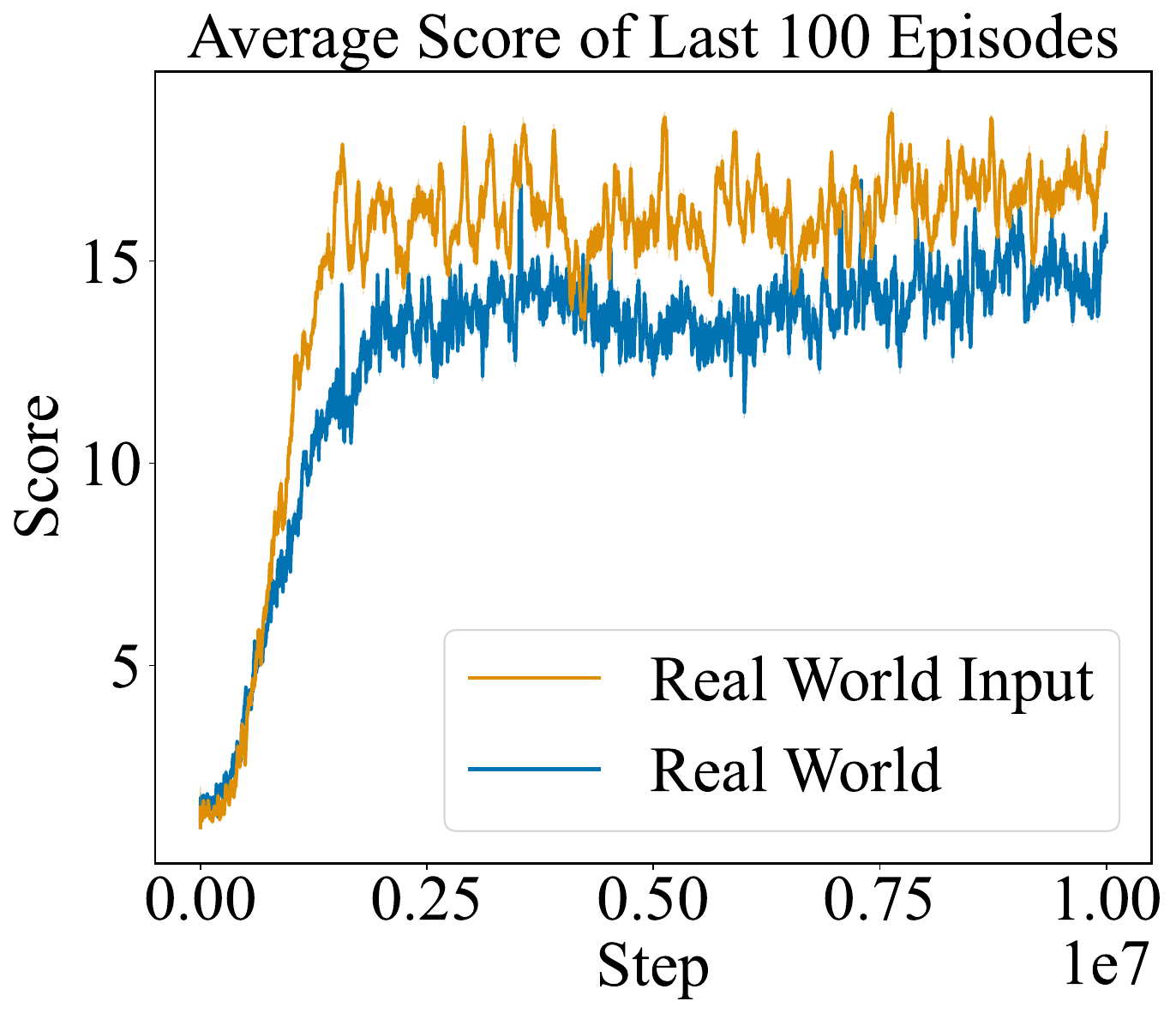}
        \caption{Learning curves of two agents across two systems}
        \label{fig:learning_curve_two_agents_two_systems}
    \end{subfigure}
    \caption{Training performance of three agents across three systems. It takes 0.3 days, 8.2 days and 19.1 days for these three agents to finish training for 10 million steps as shown in (\subref{fig:training_time_of_three_agents}). During training, the maximum mean scores over the last training episodes of three agents are 263.2, 18.8 and 17.1 as shown in (\subref{fig:maximum_mean_socre_last_100_episodes}). (\subref{fig:learning_curve_three_agents_three_systems}) and (\subref{fig:learning_curve_two_agents_two_systems}) are the learning curves of three agents in their respective systems. From top to bottom, the first one in (\subref{fig:learning_curve_three_agents_three_systems}) is the learning curve of the agent trained in simulation; the second one in (\subref{fig:learning_curve_three_agents_three_systems}) and the first one in (\subref{fig:learning_curve_two_agents_two_systems}) are the learning curves of the agent trained in the real-world input system; the last one in (\subref{fig:learning_curve_three_agents_three_systems}) and (\subref{fig:learning_curve_two_agents_two_systems}) are the learning curves of the agent trained in the real-world system. \textit{Note:} Sim = The training is done in simulation. R/I = The training is done in the real-world input system. R/W = The training is done in the real-world system. }
    \label{fig:learning_curve_three_system}
\end{figure}

\label{sec:eval}
\subsection{Experimental conditions and setup}

The evaluation uses Breakout as the benchmark task. Three agents are trained for 10 million steps under the standard DQN algorithm, each within one of the three experimental systems. Training performance is assessed using the mean score over the most recent 100 episodes, whereas final performance is evaluated by the interquartile mean (IQM) computed from 100 test episodes. To maintain comparability with established baselines, all implementations are derived from the Stable-Baselines3 framework, which is widely used as a reference platform for RL experiments and baseline comparisons \cite{sb3}.

\subsection{Sample efficiency is substantially lower in the presence of real-world factors compared to simulation}

Training in the real-world setup is substantially slower and cannot be readily accelerated because the interaction rate is constrained by physical hardware limits. Specifically, the game runs at 60 FPS, the camera is limited to 120 FPS, and the hardware-emulated keyboard has a key-press duration of 116.24 ms. Under these constraints, training for 10 million steps requires approximately 8.2 days in the real-world input system and 19.1 days in the real-world environment, whereas the same training budget is completed in about 0.3 days in simulation. The corresponding results are presented in Fig.~\ref{fig:learning_curve_three_system}(\subref{fig:training_time_of_three_agents}).

\begin{table}[!t]
\begin{center}
\caption{Human normalized scores of three agents across three systems}
\label{3sysdata}
\begin{tabular}{| c | c | c | c | c |}
\hline
Agent & Description & 95\% CI LB & PE & 95\% CI UB \\
\hline
simulation & Median & 11.08681 & 11.60764 & 11.60764 \\
\cline{2-5}
trained & IQM & 11.32222 & 11.60764 & 12.01389 \\ 
\cline{2-5}
agent & Mean & 10.92083 & 11.32049 & 11.70382 \\
\cline{2-5}
 & Optimality Gap & 0 & 0 & 0 \\
\hline
real-world & Median & 0.60069 & 0.67014 & 0.70486 \\
\cline{2-5}
input & IQM & 0.60278 & 0.65139 & 0.70417 \\ 
\cline{2-5}
trained  & Mean & 0.64201 & 0.69097 & 0.74272 \\ 
\cline{2-5}
agent & Optimality Gap & 0.29514 & 0.33490 & 0.37431 \\
\hline
real-world & Median & 0.46181 & 0.49653 & 0.53125 \\
\cline{2-5}
 system & IQM & 0.45417 & 0.49444 & 0.52986 \\ 
\cline{2-5}
trained & Mean & 0.45590 & 0.49028 & 0.52431 \\
\cline{2-5}
agent & Optimality Gap & 0.47569 & 0.50972 & 0.54410 \\
\hline
\end{tabular}
\end{center}
{\footnotesize Note: The table presents the point estimates (PE) and the corresponding lower bounds (95\% CI LB) and upper bounds (95\% CI UB) of the 95\% confidence interval of the mean, median, IQM, and optimality gap of human normalized scores achieved by three agents in their respective systems over 100 runs.}
\end{table}

\begin{figure}[htbp]
    \centering
    \begin{subfigure}{0.48\columnwidth}
        \centering
        \includegraphics[width=\linewidth]{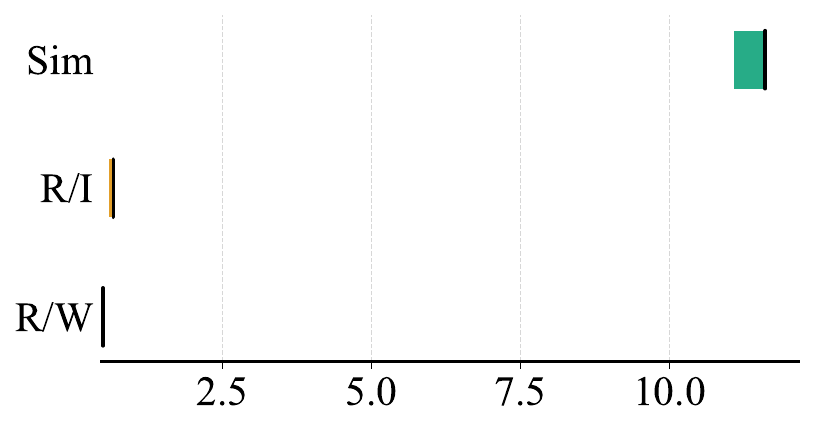}
        \caption{Median human normalized scores}
        \label{fig:three_systems_median}
    \end{subfigure}
    \hfill
    \begin{subfigure}{0.48\columnwidth}
        \centering
        \includegraphics[width=\linewidth]{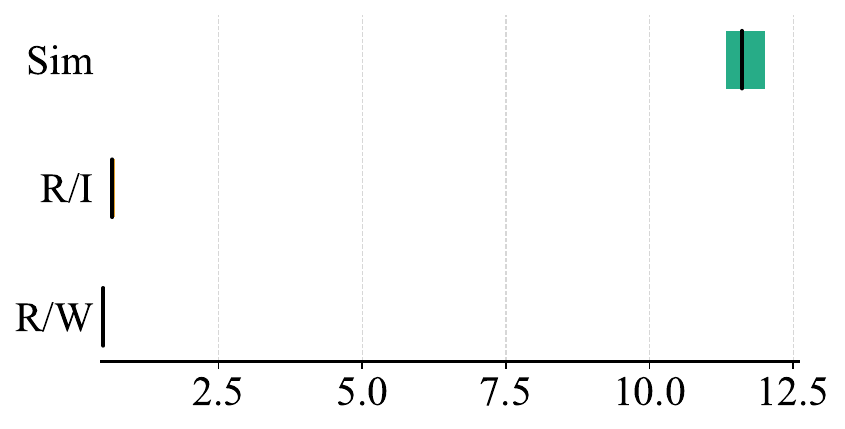}
        \caption{IQM human normalized scores}
        \label{fig:three_systems_iqm}
    \end{subfigure}
    \hfill
    \begin{subfigure}{0.48\columnwidth}
        \centering
        \includegraphics[width=\linewidth]{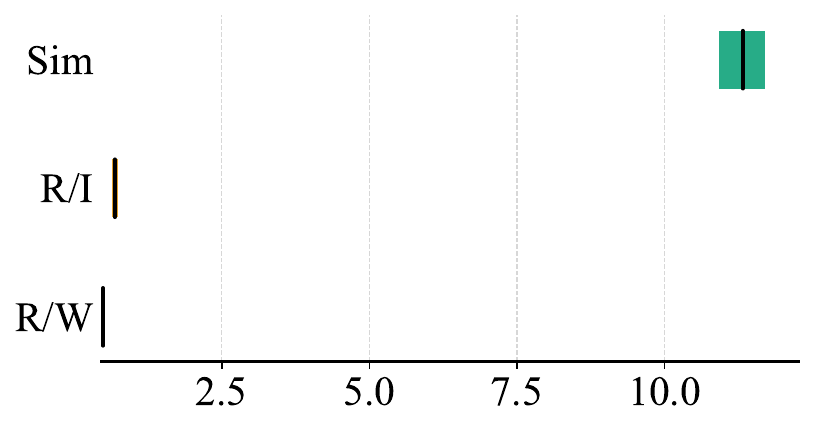}
        \caption{Mean human normalized scores}
        \label{fig:three_systems_mean}
    \end{subfigure}
    \hfill
    \begin{subfigure}{0.48\columnwidth}
       \centering
        \includegraphics[width=\linewidth]{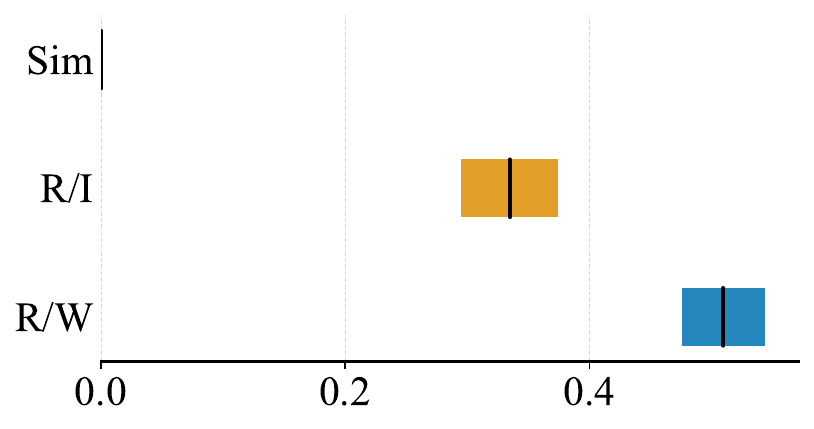}
        \caption{Optimality gap}
        \label{fig:three_systems_optimality_gap}
    \end{subfigure}
    \caption{The performance of three agents across three systems. These three agents are the agents trained in their respective systems. \textit{Note:} Sim = The agent is trained and evaluated in simulation. R/I = The agent is trained and evaluated in the real-world input system. R/W = The agent is trained and evaluated in the real-world system.}
\label{fig:3sysdiagram}
\end{figure}

\begin{figure}[htbp]
    \centering
    \begin{subfigure}{0.48\columnwidth}
        \centering
        \includegraphics[width=\linewidth]{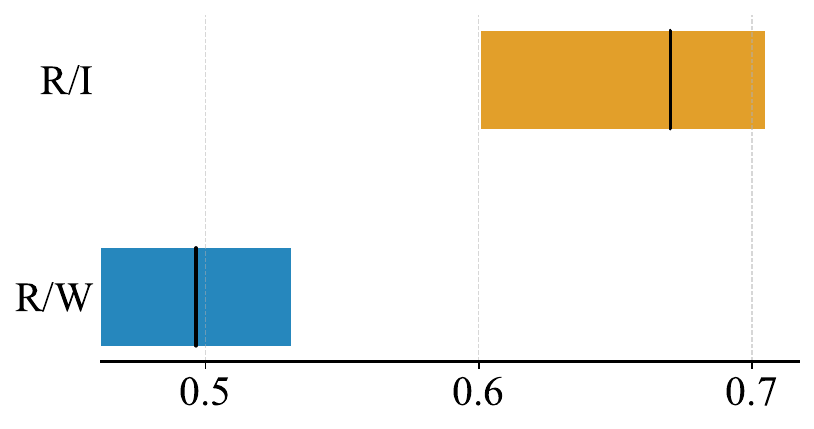}
        \caption{Median human normalized scores}
        \label{fig:two_systems_median}
    \end{subfigure}
    \hfill
    \begin{subfigure}{0.48\columnwidth}
        \centering
        \includegraphics[width=\linewidth]{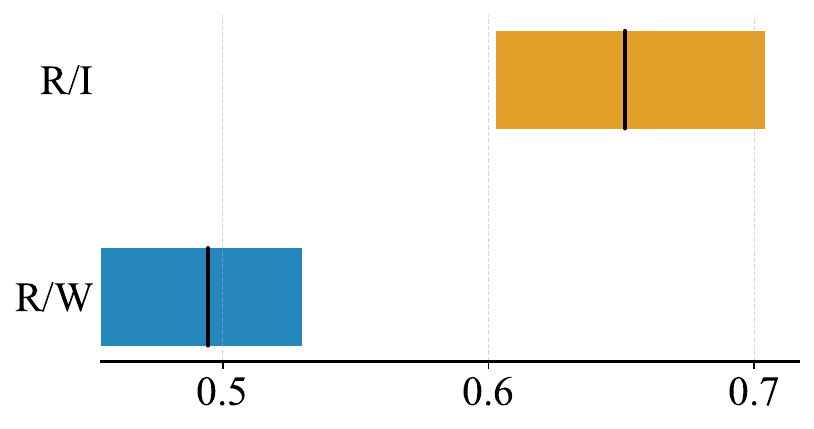}
        \caption{IQM human normalized scores}
        \label{fig:two_systems_iqm}
    \end{subfigure}
    \hfill
    \begin{subfigure}{0.48\columnwidth}
        \centering
        \includegraphics[width=\linewidth]{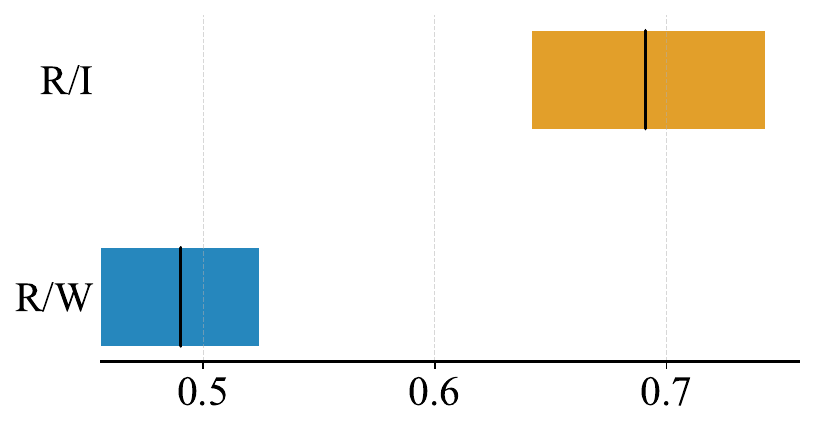}
        \caption{Mean human normalized scores}
        \label{fig:two_systems_mean}
    \end{subfigure}
    \hfill
    \begin{subfigure}{0.48\columnwidth}
       \centering
        \includegraphics[width=\linewidth]{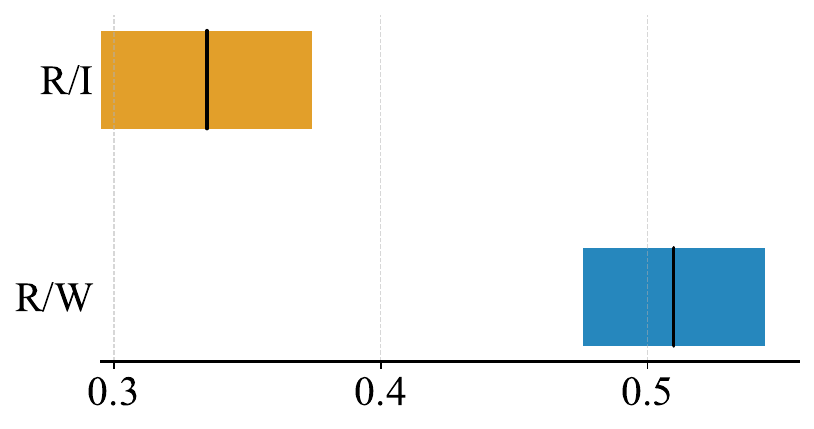}
        \caption{Optimality gap}
        \label{fig:two_systems_optimality_gap}
    \end{subfigure}
    \caption{The performance of two agents. These two agents are the agents trained in their respective systems. \textit{Note:} R/I = The agent is trained and evaluated in the real-world input system. R/W = The agent is trained and evaluated in the real-world system.}
\label{fig:2sysdiagram}
\end{figure}

\subsection{Training performance is significantly lower in environments with real-world factors than in simulation}

The agent trained in the simulation environment for 10 million steps achieves the strongest learning performance, reaching a mean score above 220 and the maximum mean score of 263.2 over the most recent 100 episodes, as shown by the highest curve in Fig.~\ref{fig:learning_curve_three_system} (\subref{fig:learning_curve_three_agents_three_systems}) and the left column in Fig.~\ref{fig:learning_curve_three_system} (\subref{fig:maximum_mean_socre_last_100_episodes}). When the observation pipeline is replaced with camera-captured input, the agent trained in the real-world input system exhibits a substantially lower learning curve, which is the middle curve as shown in Fig.~\ref{fig:learning_curve_three_system} (\subref{fig:learning_curve_three_agents_three_systems}) or the upper curve as shown in Fig.~\ref{fig:learning_curve_three_system} (\subref{fig:learning_curve_two_agents_two_systems}), with a final mean score of approximately 18 as shown in Fig.~\ref{fig:learning_curve_three_system} (\subref{fig:learning_curve_three_agents_three_systems}) and (\subref{fig:learning_curve_two_agents_two_systems}) and the maximum mean score of 18.8 as shown in the middle column in Fig.~\ref{fig:learning_curve_three_system} (\subref{fig:maximum_mean_socre_last_100_episodes}), corresponding to roughly one-fifteenth of the simulation-trained result. The lowest performance is observed for the agent trained directly in the real-world system. The final mean score is approximately 15 as shown in Fig.~\ref{fig:learning_curve_three_system} (\subref{fig:learning_curve_three_agents_three_systems}) and Fig.~\ref{fig:learning_curve_three_system} (\subref{fig:learning_curve_two_agents_two_systems}) and the maximum mean score is 17.1 as shown in Fig.~\ref{fig:learning_curve_three_system} (\subref{fig:maximum_mean_socre_last_100_episodes}) after the same training budget. The separation among the three learning curves indicates a pronounced degradation in training performance as progressively more real-world factors are introduced into the interaction loop.

\subsection{Sim-to-Real algorithmic robustness is observable but limited}
Following training, three agents are available for evaluation: one trained in the simulation environment, one trained in the real-world input system, and one trained in the full real-world system. Their final performance is assessed within their respective operating environments. The results in Fig.~\ref{fig:3sysdiagram}-\ref{fig:2sysdiagram} and Table~\ref{3sysdata} indicate substantial performance disparities among the three agents, both qualitatively and quantitatively. After 10 million training steps, neither of the agents trained under real-world interaction conditions reaches human-level performance, defined by an HNS of 1.0. By contrast, the simulation-trained agent readily attains performance far above the human baseline within the simulation environment under the same training budget. Specifically, the simulation-trained agent reaches an IQM HNS of approximately 11, whereas the agent trained in the real-world input system attains an IQM HNS of about 0.65, and the agent trained in the full real-world system achieves a lower IQM HNS of approximately 0.49. The preceding results provide evidence regarding Sim-to-Real algorithmic robustness as these agents' performance is stronger than that of a random agent with an HNS of 0. Although the agents' performance in environments with real-world factors is substantially inferior to that achieved in simulation, the RL algorithm remains operational in these environments. Therefore, the execution of the RL algorithm is not constrained by the physical limitations of the game platform, sensing pipeline, and actuation mechanism.

\begin{table}[!t]
\begin{center}
\caption{Human normalized scores of two agents in the real-world input system}
\label{realinputdata}
\begin{tabular}{| c | c | c | c | c |}
\hline
Agent & Description & 95\% CI LB & PE & 95\% CI UB \\
\hline
simulation & Median & -0.02431 & 0.01042 & 0.01042 \\
\cline{2-5}
trained & IQM & -0.00694 & 0.00625 & 0.02361 \\ 
\cline{2-5}
agent & Mean & 0.01528 & 0.03160 & 0.04965 \\
\cline{2-5}
 & Optimality Gap & 0.95035 & 0.96840 & 0.98472 \\
\hline 
real-world & Median & 0.60069 & 0.67014 & 0.70486 \\
\cline{2-5}
input & IQM & 0.60278 & 0.65139 & 0.70417 \\ 
\cline{2-5}
trained  & Mean & 0.64201 & 0.69097 & 0.74272 \\ 
\cline{2-5}
agent & Optimality Gap & 0.29514 & 0.33490 & 0.37431 \\
\hline
\end{tabular}
\end{center}
{\footnotesize Note: The table presents the point estimates (PE) and the corresponding lower bounds (95\% CI LB) and upper bounds (95\% CI UB) of the 95\% confidence interval of the mean, median, IQM, and optimality gap of human normalized scores achieved by two agents in the real-world input system over 100 runs.}
\end{table}

\begin{figure}[htbp]
    \centering
    \begin{subfigure}{0.48\columnwidth}
        \centering
        \includegraphics[width=\linewidth]{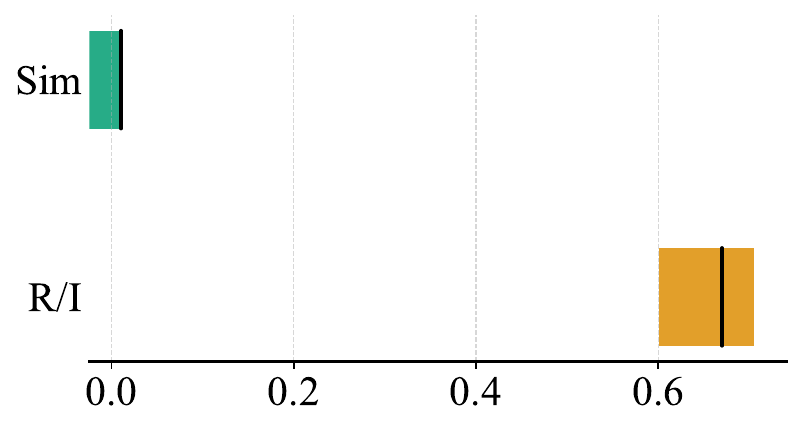}
        \caption{Median human normalized scores}
        \label{fig:real_input_system_median}
    \end{subfigure}
    \hfill
    \begin{subfigure}{0.48\columnwidth}
        \centering
        \includegraphics[width=\linewidth]{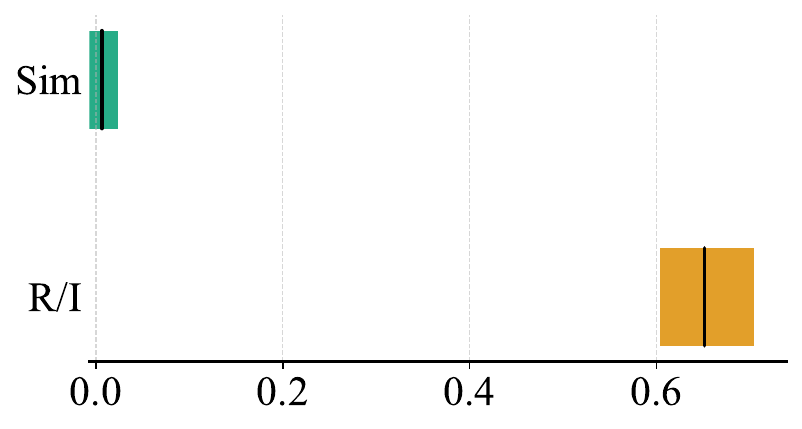}
        \caption{IQM human normalized scores}
        \label{fig:real_input_system_iqm}
    \end{subfigure}
    \hfill
    \begin{subfigure}{0.48\columnwidth}
        \centering
        \includegraphics[width=\linewidth]{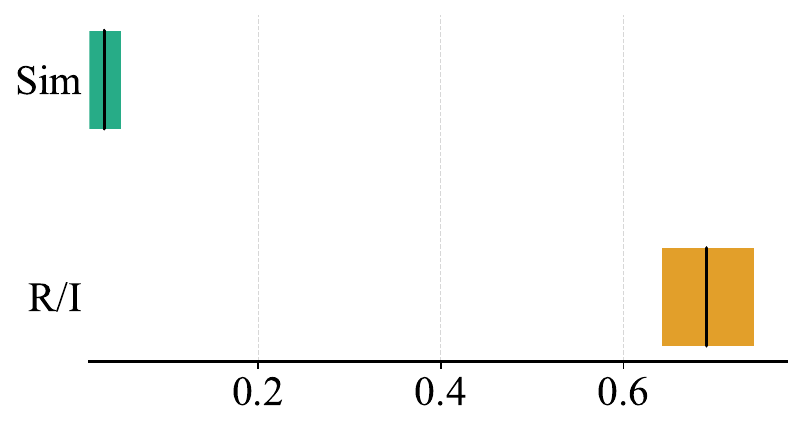}
        \caption{Mean human normalized scores}
        \label{fig:real_input_system_mean}
    \end{subfigure}
    \hfill
    \begin{subfigure}{0.48\columnwidth}
       \centering
        \includegraphics[width=\linewidth]{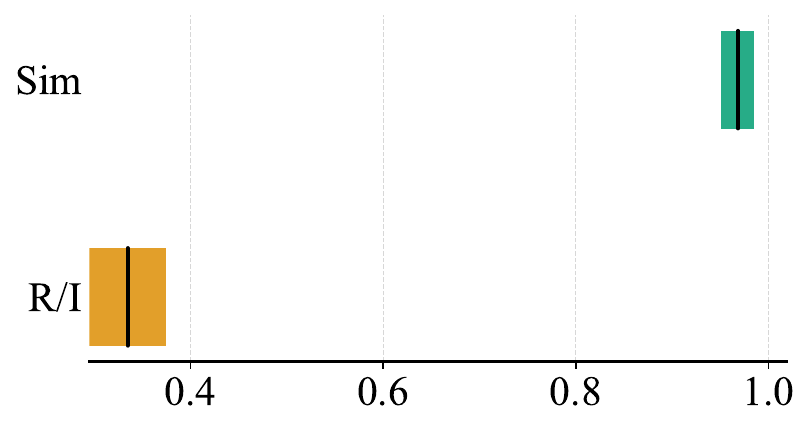}
        \caption{Optimality gap}
        \label{fig:real_input_system_optimality_gap}
    \end{subfigure}
    \caption{The performance of two agents in the real-world input system. These two agents are the agent trained in simulation and the agent trained in the real-world input system. \textit{Note:} Sim = The agent is trained in simulation and evaluated in the real-world input system. R/I = The agent is trained and evaluated in the real-world input system.}
\label{fig:realinputdiagram}
\end{figure}

\begin{table}
\begin{center}
\caption{Human normalized scores of two agents in the real-world system}
\label{realworlddata}
\begin{tabular}{| c | c | c | c | c |}
\hline
Agent & Description & 95\% CI LB & PE & 95\% CI UB \\
\hline
simulation & Median & 0.01042 & 0.01042 & 0.02778 \\
\cline{2-5}
trained & IQM & 0.01319 & 0.02083 & 0.02778 \\ 
\cline{2-5}
agent & Mean & 0.00660 & 0.01493 & 0.02292 \\
\cline{2-5}
 & Optimality Gap & 0.97708 & 0.98507 & 0.99340 \\
\hline 
real-world & Median & 0.46181 & 0.49653 & 0.53125 \\
\cline{2-5}
 system & IQM & 0.45417 & 0.49444 & 0.52986 \\ 
\cline{2-5}
trained & Mean & 0.45590 & 0.49028 & 0.52431 \\
\cline{2-5}
agent & Optimality Gap & 0.47569 & 0.50972 & 0.54410 \\
\hline
\end{tabular}
\end{center}
{\footnotesize Note: The table presents the point estimates (PE) and the corresponding lower bounds (95\% CI LB) and upper bounds (95\% CI UB) of the 95\% confidence interval of the mean, median, IQM, and optimality gap of human normalized scores achieved by two agents in the real-world system over 100 runs.}
\end{table}

\begin{figure}[htbp]
    \centering
    \begin{subfigure}{0.48\columnwidth}
        \centering
        \includegraphics[width=\linewidth]{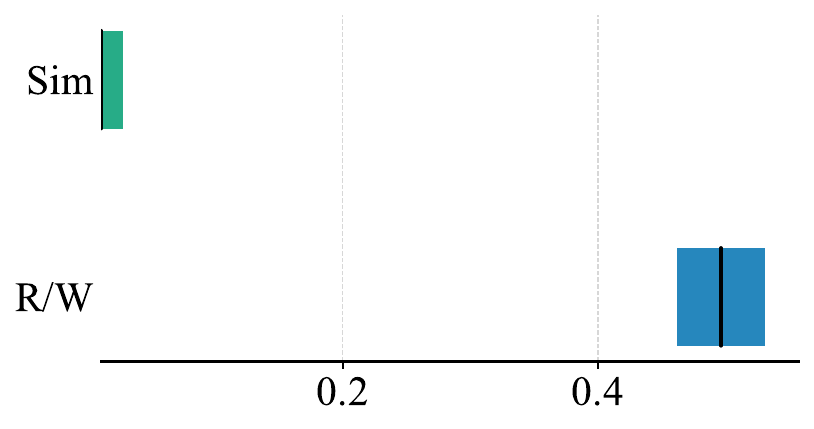}
        \caption{Median human normalized scores}
        \label{fig:real_world_system_median}
    \end{subfigure}
    \hfill
    \begin{subfigure}{0.48\columnwidth}
        \centering
        \includegraphics[width=\linewidth]{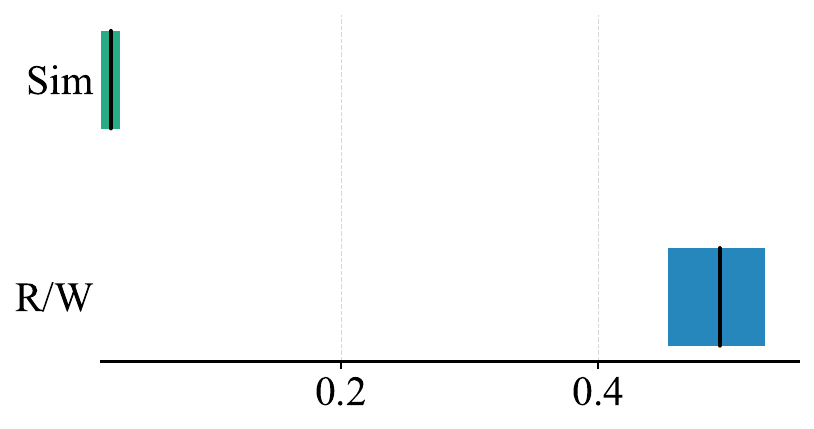}
        \caption{IQM human normalized scores}
        \label{fig:real_world_system_iqm}
    \end{subfigure}
    \hfill
    \begin{subfigure}{0.48\columnwidth}
        \centering
        \includegraphics[width=\linewidth]{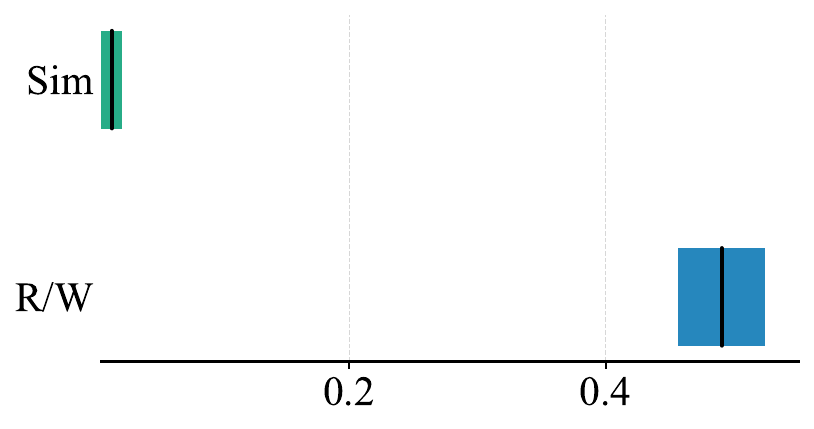}
        \caption{Mean human normalized scores}
        \label{fig:real_world_system_mean}
    \end{subfigure}
    \hfill
    \begin{subfigure}{0.48\columnwidth}
       \centering
        \includegraphics[width=\linewidth]{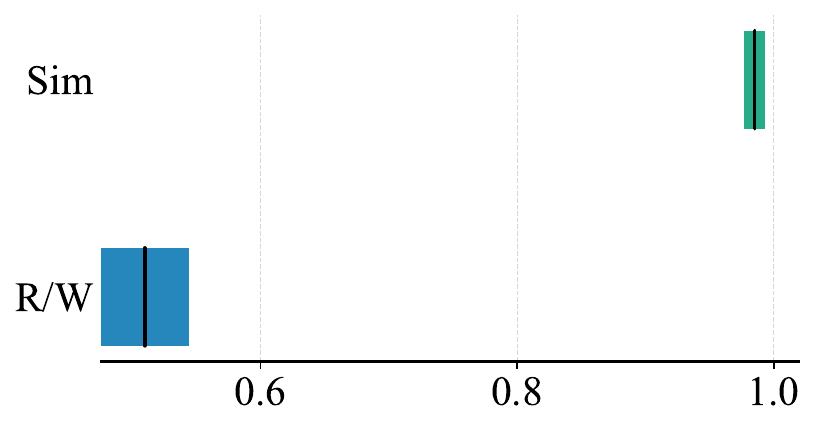}
        \caption{Optimality gap}
        \label{fig:real_world_system_optimality_gap}
    \end{subfigure}
    \caption{The performance of two agents in the real-world system. These two agents are the agent trained in simulation and the agent trained in the real-world system. \textit{Note:} Sim = The agent is trained in simulation and evaluated in the real-world system. R/W = The agent is trained and evaluated in the real-world system.}
\label{fig:realworlddiagram}
\end{figure}

\subsection{Sim-to-Real gap is severe}

To evaluate cross-environment generalization, the agent trained in the simulation environment is tested in the real-world input system, and the performance is compared with that of the agent trained in the real-world input system in Fig.~\ref{fig:realinputdiagram} and Table~\ref{realinputdata}. Subsequently, the simulation-trained agent is evaluated in the real-world system, with the corresponding results presented in Fig.~\ref{fig:realworlddiagram} and Table~\ref{realworlddata}. The simulation-trained agent achieves HNS values below 0.03 across both real-world related systems, indicating a near-complete failure to cope with real-world factors. By contrast, the agent trained in the real-world input system attains an HNS of about 0.65, and the agent trained directly in the real-world system reaches an HNS of about 0.49, indicating that operation in the real-world system is feasible, although performance remains low. These findings demonstrate a severe Sim-to-Real gap and indicate that the simulation-trained agent does not transfer effectively to systems involving real-world factors.

\begin{figure}[!t]
\centerline{\includegraphics[width=\linewidth]{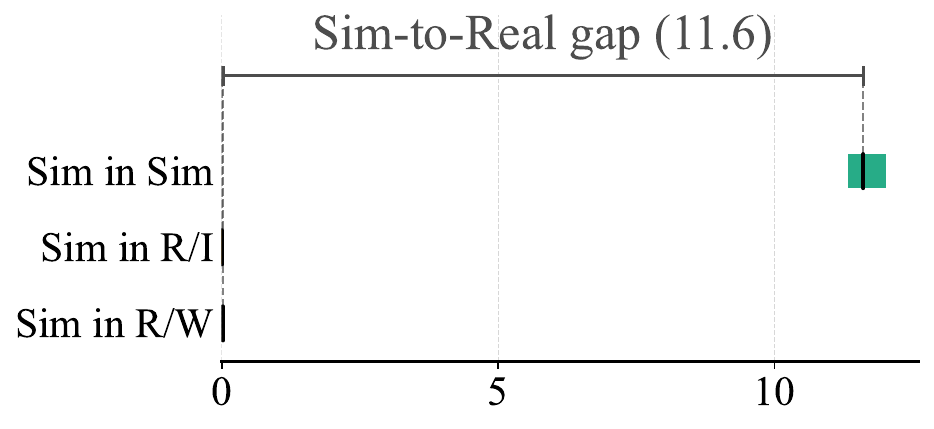}}
\caption{The Sim-to-Real gap. The difference between the performances of the simulation-trained agent in three systems. \textit{Note:} Sim in Sim = The agent is trained and evaluated in simulation. Sim in R/I = The agent is trained in simulation and evaluated in the real-world input system. Sim in R/W = The agent is trained in simulation and evaluated in the real-world system.}
\label{sim2realgap}
\end{figure}

\begin{table}[!t]
\begin{center}
\caption{Interquartile Mean (IQM) of human normalized scores of the simulation-trained agent in three systems}
\label{simin3sys}
\begin{tabular}{| c | c | c | c | c |}
\hline
Data & System & 95\% CI LB & PE & 95\% CI UB \\
\hline
IQM of the & Sim & 11.32222 & 11.60764 & 12.01389 \\
\cline{2-5}
simulation & R/I & -0.00694 & 0.00625 & 0.02361 \\ 
\cline{2-5}
trained agent & R/W & 0.01319 & 0.02083 & 0.02778 \\
\cline{2-5}
\hline
\end{tabular}
\end{center}
{\footnotesize Note: The table presents the point estimates (PE) and the corresponding lower bounds (95\% CI LB) and upper bounds (95\% CI UB) of the 95\% confidence interval of the IQM of human normalized scores achieved by the simulation-trained agent over 100 runs in three systems. Sim = The simulation-trained agent's performance in simulation. R/I = The simulation-trained agent's performance in the real-world input system. R/W = The simulation-trained agent's performance in the real-world system.}
\end{table}

\begin{figure}[htbp]
    \centering
    \begin{subfigure}{0.27\columnwidth}
        \centering
        \includegraphics[width=\linewidth]{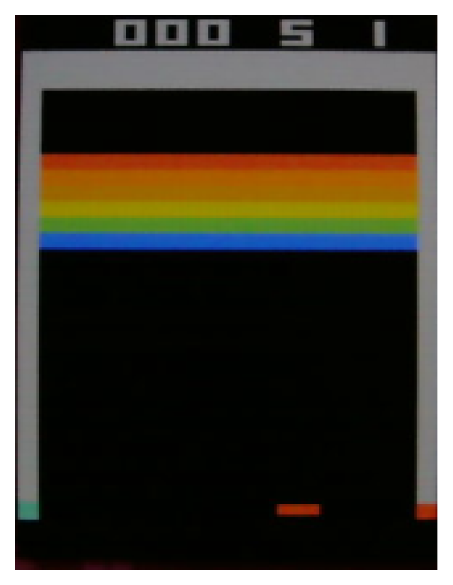}
        \caption{}
        \label{fig:cropped_image_210_160}
    \end{subfigure}
    \hfill
    \begin{subfigure}{0.27\columnwidth}
        \centering
        \includegraphics[width=\linewidth]{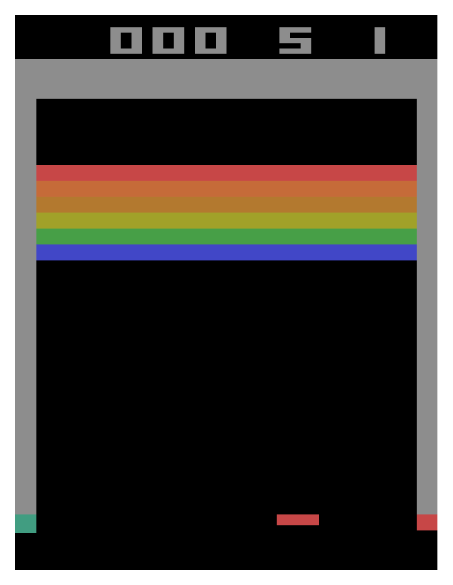}
        \caption{}
        \label{fig:env_obs_210_160}
    \end{subfigure}
    \hfill
    \begin{subfigure}{0.39\columnwidth}
        \centering
        \includegraphics[width=\linewidth]{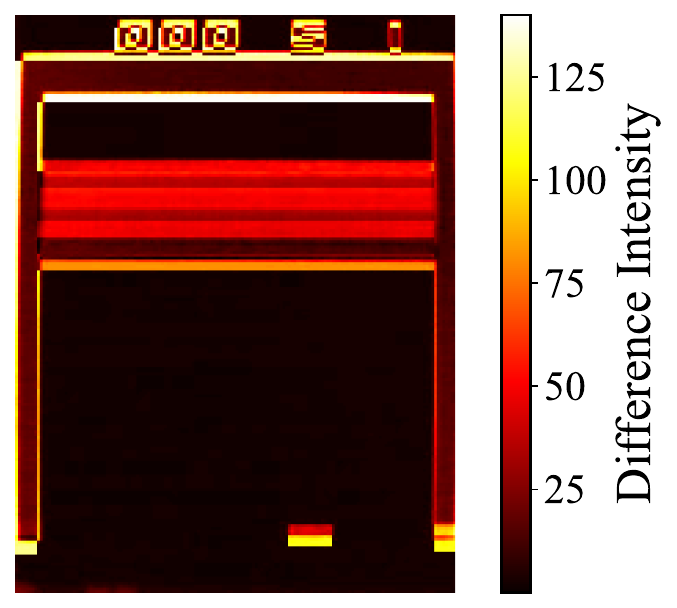}
        \caption{}
        \label{fig:difference_heatmap}
    \end{subfigure}
    \hfill
    \begin{subfigure}{0.30\columnwidth}
        \centering
        \includegraphics[width=\linewidth]{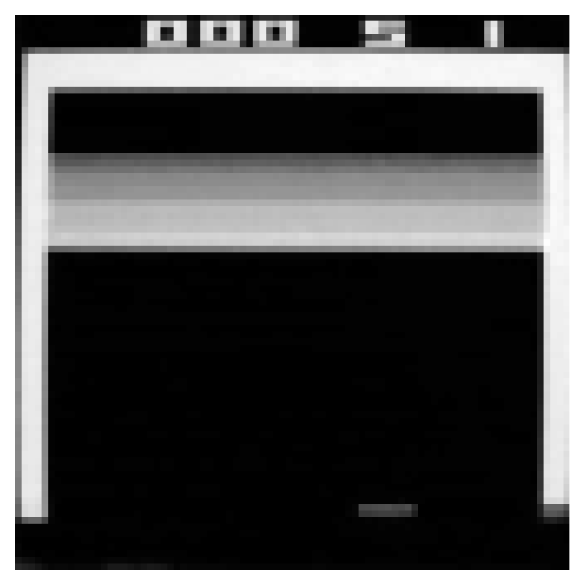}
        \caption{}
        \label{fig:cropped_resized_gray}
    \end{subfigure}
    \hfill
    \begin{subfigure}{0.30\columnwidth}
        \centering
        \includegraphics[width=\linewidth]{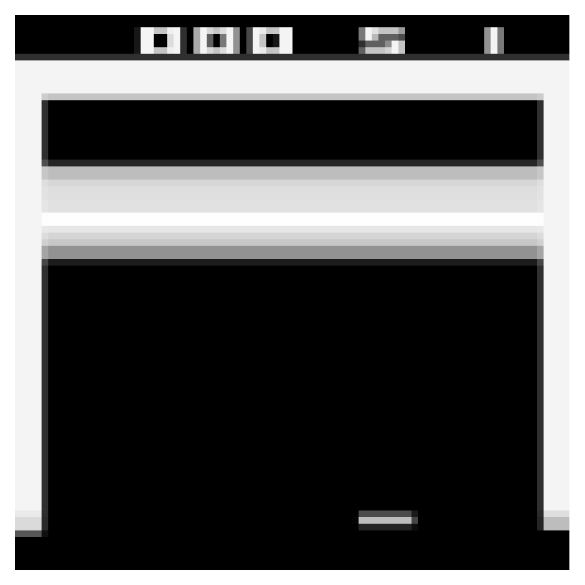}
        \caption{}
        \label{fig:obs_resized_gray}
    \end{subfigure}
    \hfill
    \begin{subfigure}{0.35\columnwidth}
        \centering
        \includegraphics[width=\linewidth]{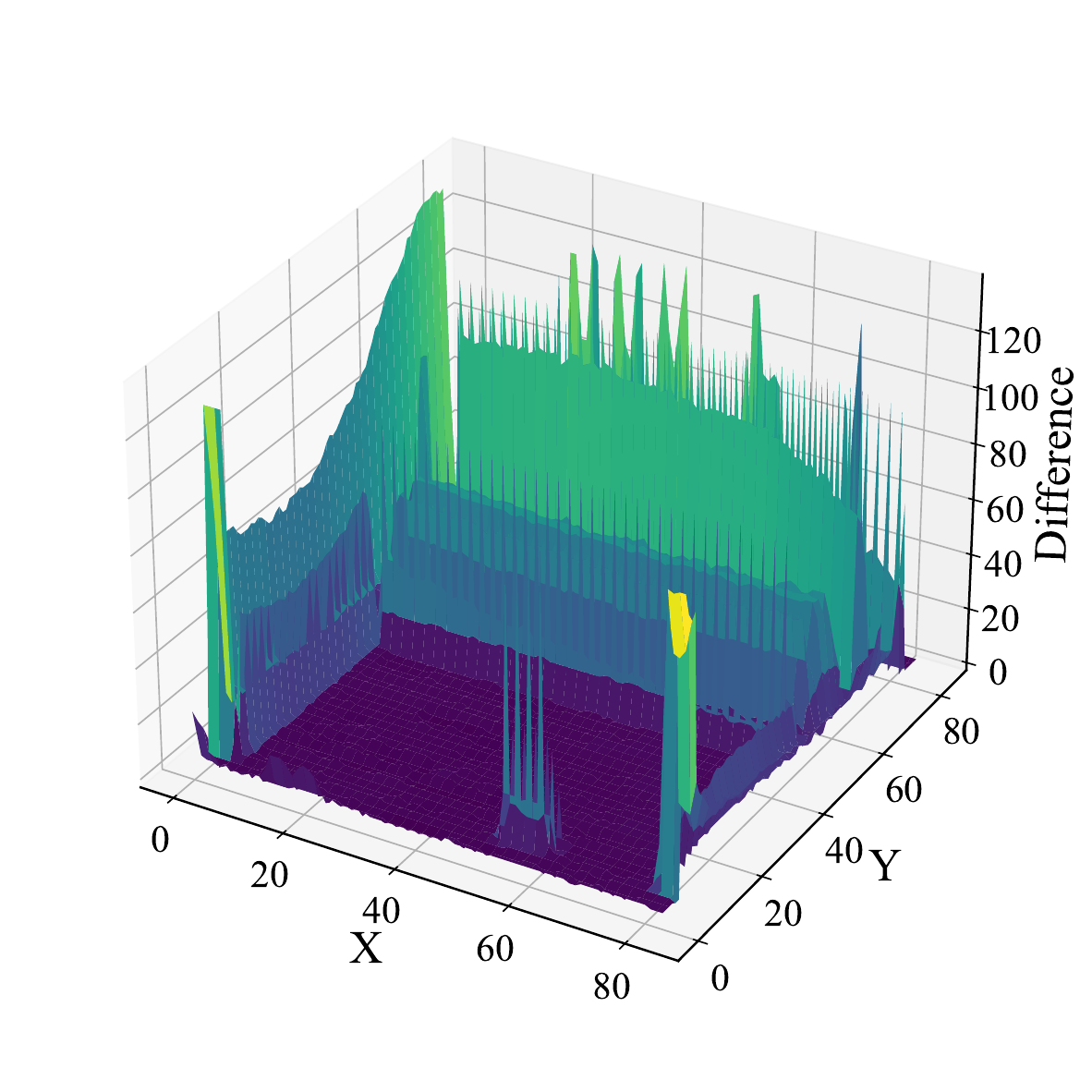}
        \caption{}
        \label{fig:3d_surface_plot}
    \end{subfigure}
    \hfill
    \begin{subfigure}{0.32\columnwidth}
        \centering
        \includegraphics[width=\linewidth]{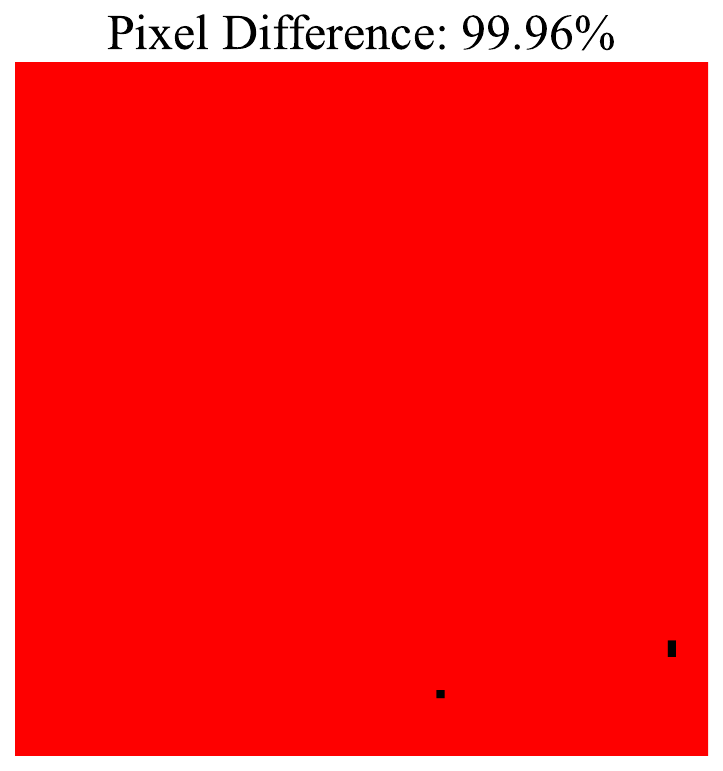}
        \caption{}
        \label{fig:threshold_0}
    \end{subfigure}
    \hfill
    \begin{subfigure}{0.32\columnwidth}
        \centering
        \includegraphics[width=\linewidth]{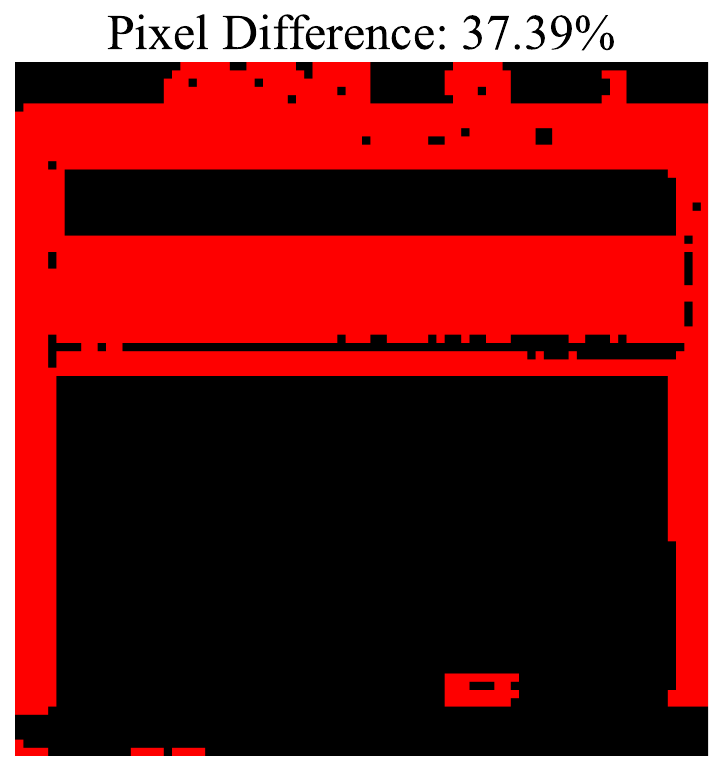}
        \caption{}
        \label{fig:threshold_10}
    \end{subfigure}
    \hfill
    \begin{subfigure}{0.32\columnwidth}
       \centering
        \includegraphics[width=\linewidth]{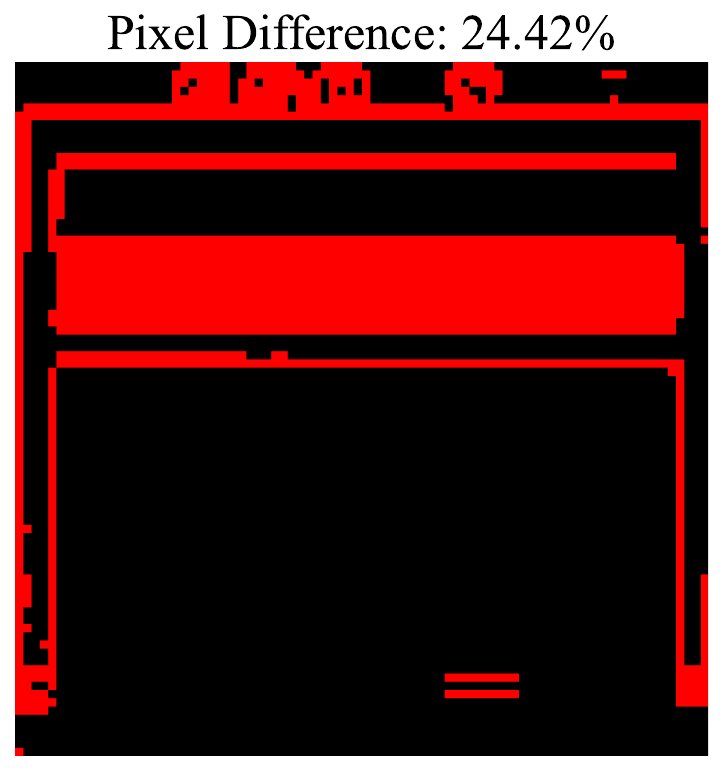}
        \caption{}
        \label{fig:threshold_20}
    \end{subfigure}
    \caption{The difference between the cropped field of view of the camera input and the observation output of the game environment. (\subref{fig:cropped_image_210_160}) Cropped field of view of the camera input, (\subref{fig:env_obs_210_160}) Simulation environment observation output; (\subref{fig:difference_heatmap}) Image difference heatmap between the two pictures in (\subref{fig:cropped_image_210_160}) and (\subref{fig:env_obs_210_160}); (\subref{fig:cropped_resized_gray}) Resized 84$\times$84 gray image of the cropped image; (\subref{fig:obs_resized_gray}) Resized 84$\times$84 gray image of the observation; (\subref{fig:3d_surface_plot}) 3D surface plot of pixel difference between the two pictures in (\subref{fig:cropped_resized_gray}) and (\subref{fig:obs_resized_gray}); (\subref{fig:threshold_0}) Pixel difference between (\subref{fig:cropped_resized_gray}) and (\subref{fig:obs_resized_gray}) when threshold = 0; (\subref{fig:threshold_10}) Pixel difference between (\subref{fig:cropped_resized_gray}) and (\subref{fig:obs_resized_gray}) when threshold = 10; (\subref{fig:threshold_20}) Pixel difference between (\subref{fig:cropped_resized_gray}) and (\subref{fig:obs_resized_gray}) when threshold = 20.}
\label{fig:difference}
\end{figure}

As shown in Fig.~\ref{sim2realgap} and based on the data in Table~\ref{simin3sys}, the Sim-to-Real gap is calculated as 11.6. Relative to the human baseline, the simulation-trained agent's performance declines by approximately 1160\%, dropping from 1061\% above human level in simulation to 99\% below human level under real-world conditions. To examine the source of this discrepancy, the cropped camera-based observation is compared with the original game screen, that is, the observation provided by the simulation environment; the comparison is illustrated in Fig.~\ref{fig:difference}. This comparison reveals modest visual degradations in the camera-derived input, including distortion, reduced brightness, color deviation, and blurring. Although these differences are not substantial for human perception and do not prevent a human player from performing the task, they are sufficient to impair the behavior of the simulation-trained agent. These findings indicate that the agent trained in simulation lacks the generalization capability required for robust operation in real-world environments.

\subsection{No adaptation to minor input changes}

\begin{figure}[!t]
\centering
\includegraphics[width=\linewidth]{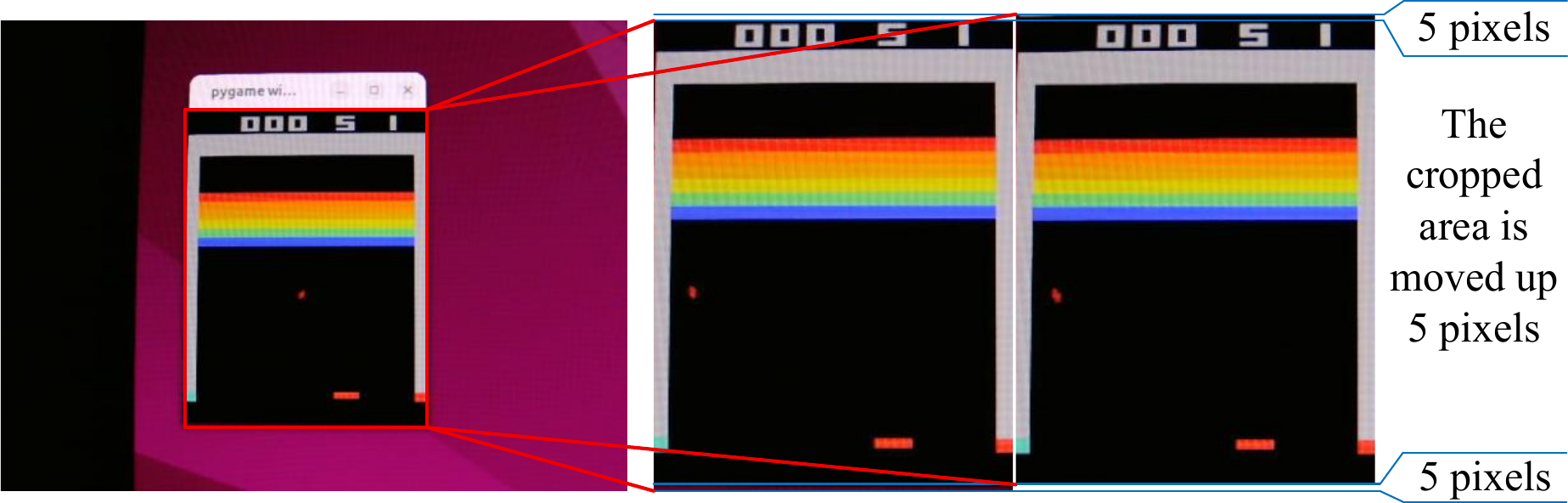}
\caption{The cropped area is moved up 5 pixels in the real-world input system.}
\label{fig:ri5pixels}
\end{figure}

\begin{figure}[htbp]
\centering
\includegraphics[width=\linewidth]{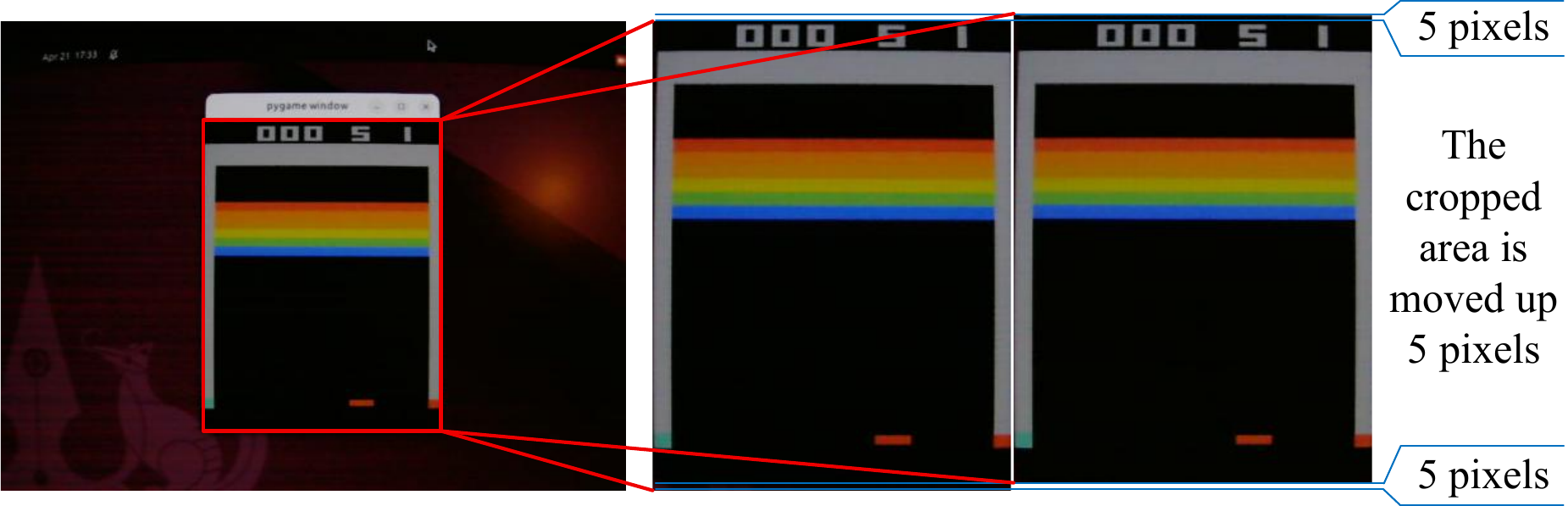}
\caption{The cropped area is moved up 5 pixels in the real-world system.}
\label{fig:rw5pixels}
\end{figure}

Even when agents are trained in the environments that incorporate real-world factors, they fail to generalize to minor variations in sensory inputs. When the cropped region of the field of view of the camera is shifted upward by 5 pixels in both the real-world input system and the real-world system, as shown in Fig.~\ref{fig:ri5pixels} and Fig.~\ref{fig:rw5pixels}, the agents trained in these systems become nonfunctional. Performance in the real-world input system decreases from 0.65 to 0.02, and the performance in the real-world system drops from 0.49 to 0.02. These results prove that all these agents lack adaptability to environmental changes, despite being trained directly on the nonstationary data from these systems.

\begin{table}
\begin{center}
\caption{Human normalized scores of the agent in the real-world input system before and after the change}
\label{realinputafterchange}
\begin{tabular}{| c | c | c | c | c |}
\hline
Agent & Description & 95\% CI LB & PE & 95\% CI UB \\
\hline
performance & Median & 0.01042 & 0.01042 & 0.01042 \\
\cline{2-5}
after & IQM & 0.00694 & 0.01736 & 0.02361 \\ 
\cline{2-5}
the change & Mean & 0.01181 & 0.01910 & 0.02708 \\
\cline{2-5}
 & Optimality Gap & 0.97292 & 0.98090 & 0.98819 \\
\hline 
performance & Median & 0.60069 & 0.67014 & 0.70486 \\
\cline{2-5}
before & IQM & 0.60278 & 0.65139 & 0.70417 \\ 
\cline{2-5}
the change  & Mean & 0.64201 & 0.69097 & 0.74272 \\ 
\cline{2-5}
 & Optimality Gap & 0.29514 & 0.33490 & 0.37431 \\
\hline
\end{tabular}
\end{center}
{\footnotesize Note: The table presents the point estimates (PE) and the corresponding lower bounds (95\% CI LB) and upper bounds (95\% CI UB) of the 95\% confidence interval of the mean, median, IQM, and optimality gap of human normalized scores achieved by the agent in the real-world input system over 100 runs.}
\end{table}

\begin{figure}[htbp]
    \centering
    \begin{subfigure}{0.48\columnwidth}
        \centering
        \includegraphics[width=\linewidth]{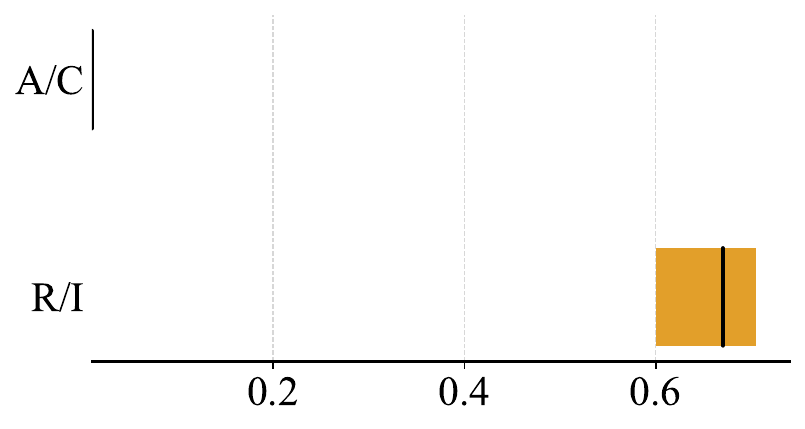}
        \caption{Median human normalized scores}
        \label{fig:real_world_input_change_median}
    \end{subfigure}
    \hfill
    \begin{subfigure}{0.48\columnwidth}
        \centering
        \includegraphics[width=\linewidth]{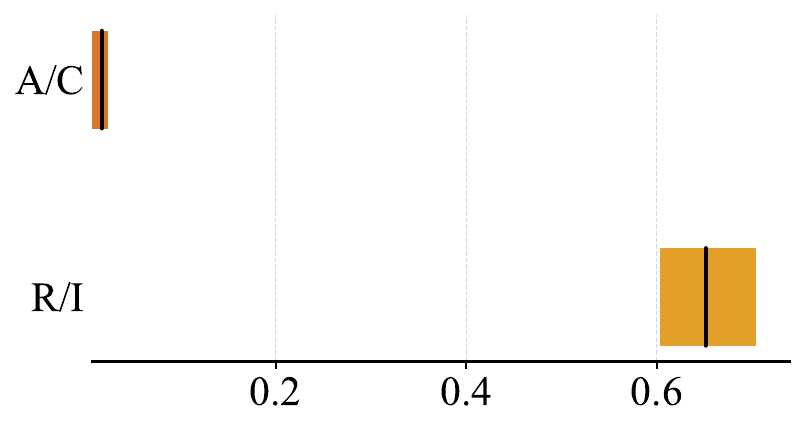}
        \caption{IQM human normalized scores}
        \label{fig:real_world_input_change_iqm}
    \end{subfigure}
    \hfill
    \begin{subfigure}{0.48\columnwidth}
        \centering
        \includegraphics[width=\linewidth]{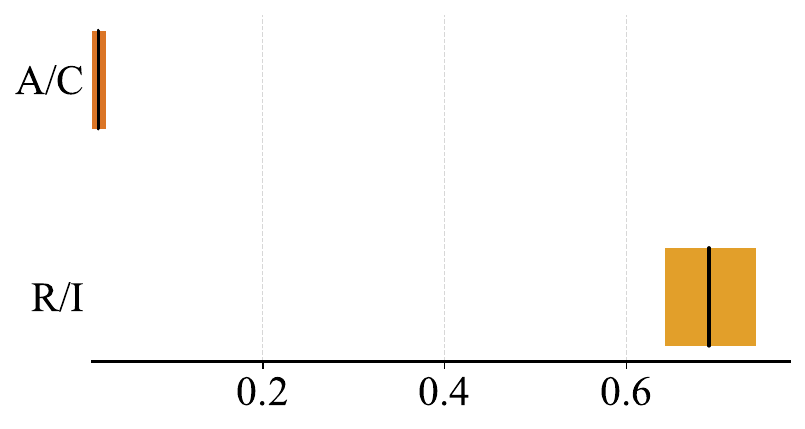}
        \caption{Mean human normalized scores}
        \label{fig:real_world_input_change_mean}
    \end{subfigure}
    \hfill
    \begin{subfigure}{0.48\columnwidth}
       \centering
        \includegraphics[width=\linewidth]{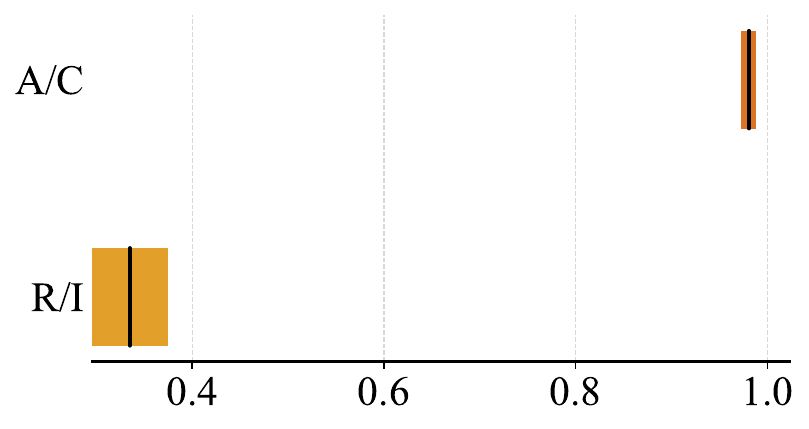}
        \caption{Optimality gap}
        \label{fig:real_world_input_change_optimality_gap}
    \end{subfigure}
    \caption{The performance of the agent in the real-world input system before and after the change. \textit{Note:} A/C = The agent is evaluated in the real-world input system after the input change. R/I = The agent is evaluated in the real-world input system before the input change.}
\label{fig:realworldinputchangediagram}
\end{figure}

\begin{table}
\begin{center}
\caption{Human normalized scores of the agent in the real-world system before and after the change}
\label{realworldafterchange}
\begin{tabular}{| c | c | c | c | c |}
\hline
Agent & Description & 95\% CI LB & PE & 95\% CI UB \\
\hline
performance & Median & 0.01042 & 0.01042 & 0.04514 \\
\cline{2-5}
after & IQM & 0.01597 & 0.02292 & 0.02986 \\ 
\cline{2-5}
the change & Mean & 0.02188 & 0.02917 & 0.03681 \\
\cline{2-5}
 & Optimality Gap & 0.96319 & 0.97083 & 0.97813 \\
\hline 
performance & Median & 0.46181 & 0.49653 & 0.53125 \\
\cline{2-5}
before & IQM & 0.45417 & 0.49444 & 0.52986 \\ 
\cline{2-5}
the change & Mean & 0.45590 & 0.49028 & 0.52431 \\
\cline{2-5}
 & Optimality Gap & 0.47569 & 0.50972 & 0.54410 \\
\hline
\end{tabular}
\end{center}
{\footnotesize Note: The table presents the point estimates (PE) and the corresponding lower bounds (95\% CI LB) and upper bounds (95\% CI UB) of the 95\% confidence interval of the mean, median, IQM, and optimality gap of human normalized scores achieved by the agent in the real-world system over 100 runs.}
\end{table}

\begin{figure}[htbp]
    \centering
    \begin{subfigure}{0.48\columnwidth}
        \centering
        \includegraphics[width=\linewidth]{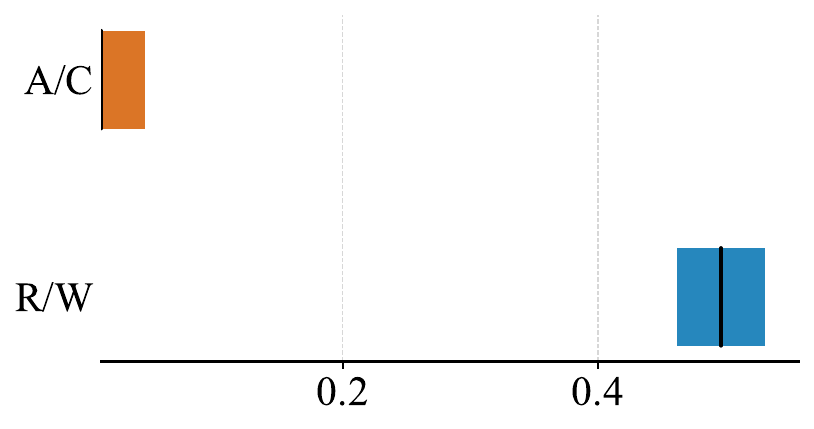}
        \caption{Median human normalized scores}
        \label{fig:real_world_input_change_median}
    \end{subfigure}
    \hfill
    \begin{subfigure}{0.48\columnwidth}
        \centering
        \includegraphics[width=\linewidth]{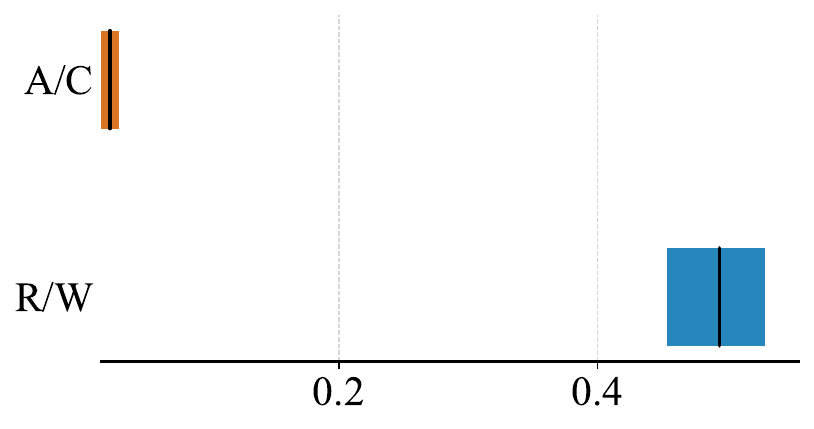}
        \caption{IQM human normalized scores}
        \label{fig:real_world_input_change_iqm}
    \end{subfigure}
    \hfill
    \begin{subfigure}{0.48\columnwidth}
        \centering
        \includegraphics[width=\linewidth]{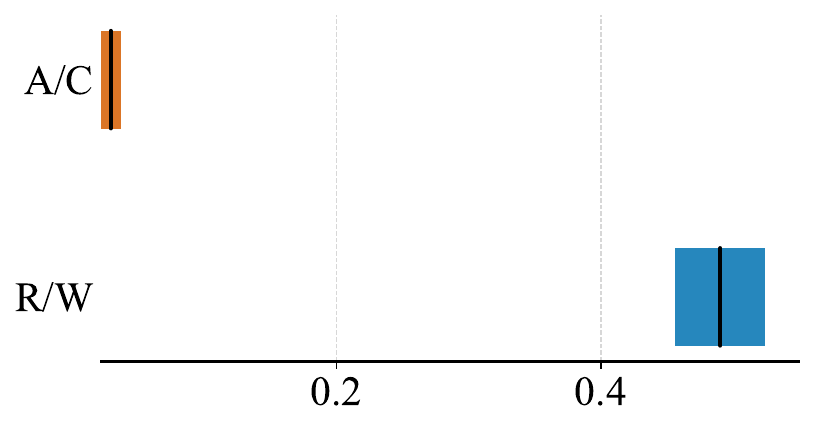}
        \caption{Mean human normalized scores}
        \label{fig:real_world_input_change_mean}
    \end{subfigure}
    \hfill
    \begin{subfigure}{0.48\columnwidth}
       \centering
        \includegraphics[width=\linewidth]{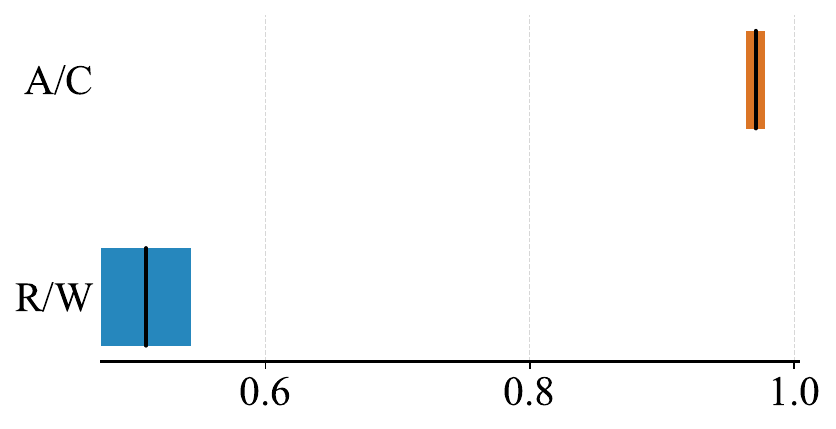}
        \caption{Optimality gap}
        \label{fig:real_world_input_change_optimality_gap}
    \end{subfigure}
    \caption{The performance of the agent in the real-world system before and after the change. \textit{Note:} A/C = The agent is evaluated in the real-world system after the input change. R/W = The agent is evaluated in the real-world system before the input change.}
\label{fig:realworldsystemchangediagram}
\end{figure}

\subsection{Sim-to-Real benchmark platform for AIoT is validated}

The evaluation results indicate that the environments incorporating real-world factors, which introduce multiple limiting and time-varying factors such as latency and noise, substantially affect both the training dynamics and final performance of RL agents. When real-world factors are present during either training or evaluation, agent performance degrades markedly. In particular, performance declines from a level well above the human baseline ($\text{HNS} > 1.0$) to a level far below human performance ($\text{HNS} < 1.0$), and in some cases approaches that of a random agent ($\text{HNS} \approx 0.0$). These findings demonstrate that the two implemented systems, which incorporate real-world factors, provide an effective testbed for evaluating simulation-trained agents. In addition, both systems can serve as practical real-world platforms for RL training. The benchmark platform is validated for the Sim-to-Real gap measurement and the Sim-to-Real algorithmic robustness assessment for RL in AIoT.

\section{Discussion}
\label{sec:dis}
\subsection{RL algorithms work in real-world applications}

The learning curves in Fig.~\ref{fig:learning_curve_three_system} indicate that the agent trained in the real-world system exhibits sustained performance improvement over the 10 million training steps. The most pronounced gain occurs within the first 1 million steps, during which the mean score increases from below 2 to approximately 12. This trend demonstrates that the conventional DQN algorithm remains effective in the implemented real-world system.

\subsection{The performance gap between agents can be identified and measured}

With respect to final performance, the disparities among agents can be identified and quantified using the IQM values obtained from 100 test episodes, as presented in Fig.~\ref{fig:3sysdiagram}, \ref{fig:realinputdiagram}, and \ref{fig:realworlddiagram} and summarized in Table~\ref{3sysdata}, \ref{realinputdata}, and \ref{realworlddata}. For example, in the real-world system, the agent trained directly in that environment achieves an HNS of approximately 0.49. This result exceeds the performance of the simulation-trained agent by about 0.47.

\subsection{Real-world system is affordable and easily accessible}

The simulation environment can be run on any computer. The real-world input system has a camera, a camera stand, and a Linux computer. The real-world system comprises a camera, a camera stand, a USB-to-UART converter with cable, a Teensy board, two USB-to-UART cables, a null modem, and two Linux computers. Excluding the computers, the total hardware cost for the benchmark platform was below USD 400 at the time of procurement, as the computers, camera, and camera stand can be reused across these systems. All required components are commercially available and can be obtained readily through online vendors. As a result, the platform enables researchers to investigate real-world RL systems under substantially reduced financial constraints. Moreover, because the control objective is limited to maximizing video-game score, the experimental setting avoids the safety risks typically associated with physical control tasks.

\subsection{It is challenging to train agents in the real-world system}

A primary challenge lies in accommodating real-world factors, including sensor noise and actuation latency, during training. Owing to these physical constraints, agents trained in real-world related systems attain performance approximately 15--20 times lower than that of the agent trained in simulation, and the simulation-trained agent also fails to perform effectively when deployed in any of the real-world related systems. These findings highlight the difficulty of developing a robust RL algorithm capable of training a generalized agent that operates reliably in both simulated and physical environments. Another major challenge is the slow pace of real-world training. Because sensing, data processing, and actuation are all subject to physical time constraints, the training process cannot be accelerated in the same manner as in simulation.

\section{Conclusion}
\label{sec:con}

The developed platform provides a practical foundation for qualitative and quantitative analysis of RL agent performance in real-world AIoT scenarios. It serves as a low-cost physical benchmark for research on RL algorithms in AIoT systems and supports both real-world training and empirical evaluation under AIoT-relevant conditions. By enabling direct measurement of agent performance, the Sim-to-Real gap and Sim-to-Real algorithmic robustness can be systematically measured and assessed, thereby facilitating rigorous evaluation of whether newly proposed methods or algorithms yield meaningful improvements. In the present study, experimental results for an agent trained in simulation reveal a substantial and measurable discrepancy between performance in simulation and performance on the proposed real-world platform. This discrepancy highlights a critical research direction for reinforcement learning in AIoT systems, namely, the development of methods that improve transferability and robustness under real-world conditions.

\section{Future Work}
\label{sec:future}
\subsection{Solve Sim-to-Real RL problems in the real-world input system first}

The experimental results indicate that real-world visual input alone has already introduced a substantial degradation in agent performance, even before actuation is incorporated to form the full real-world system. Accordingly, future work will first address RL challenges in the real-world input system before extending the investigation to the complete real-world system.

\subsection{Shorten the training time in a real-world environment}

Training an agent for 10 million steps requires more than 8 days in the real-world input environment and approximately 19 days in the full real-world environment. Reducing training time in real-world settings is therefore a necessary objective. Because real-world reinforcement learning cannot be arbitrarily accelerated due to hardware-imposed constraints, future work will investigate both existing and newly developed methods to improve training efficiency. One promising direction is parallel learning, in which multiple real-world environments are operated concurrently. Under this setting, parallel RL algorithms such as A3C \cite{pmlr-v48-mniha16} and PPO \cite{schulman2017proximalpolicyoptimizationalgorithms} can aggregate experience across several real-world systems, thereby increasing data collection efficiency and shortening overall training time.

\section{Platform Availability}
\label{sec:platform}
The benchmark platform software and its associated hardware implementation instructions are available upon request from the author. Because reinforcement learning agents trained on one hardware platform typically exhibit limited transferability to other platform instances, experimental reproducibility is achieved by reconstructing the benchmark hardware platform and conducting training and evaluation locally. Researchers can therefore replicate the benchmark by following the provided implementation guidelines to build the proposed platform and perform agent training and testing on the platform they construct.

\section*{Acknowledgment}
Professor Albert Zomaya would like to acknowledge the support of the Australian Research Council Research Hub for Future Digital Manufacturing (IH230100013).

\newpage

 




\vfill

\end{document}